\documentclass{article}

\usepackage{microtype}
\usepackage{graphicx}
\usepackage{subfigure}
\usepackage{booktabs} 

\usepackage{hyperref}

\usepackage[accepted]{icml2023}

\usepackage{amsmath}
\usepackage{amssymb}
\usepackage{mathtools}
\usepackage{amsthm}

\usepackage[capitalize,noabbrev]{cleveref}

\theoremstyle{plain}

\theoremstyle{definition}

\theoremstyle{remark}

\usepackage[textsize=tiny]{todonotes}

\usepackage{amsmath,amsfonts,bm}

\def\eqref#1{equation~\ref{#1}}

\def\1{\bm{1}}

\DeclareMathAlphabet{\mathsfit}{\encodingdefault}{\sfdefault}{m}{sl}
\SetMathAlphabet{\mathsfit}{bold}{\encodingdefault}{\sfdefault}{bx}{n}

\DeclareMathOperator*{\argmax}{arg\,max}

\usepackage{wrapfig}
\usepackage{bm}
\usepackage{multirow,makecell}
\usepackage{enumitem}
\usepackage{xcolor}
\usepackage{color, colortbl}
\usepackage[most]{tcolorbox}
\definecolor{Gray}{gray}{0.9}

\newcommand{\alg}{\textsc{PromptBoosting}}

\usepackage{enumitem}

\icmltitlerunning{PromptBoosting: Black-Box Text Classification with Ten Forward Passes}

\begin{document}

\twocolumn[
\icmltitle{PromptBoosting: Black-Box Text Classification with Ten Forward Passes}



\icmlsetsymbol{equal}{*}

\begin{icmlauthorlist}
\icmlauthor{Bairu Hou}{ucsb}
\icmlauthor{Joe O’Connor}{ucla}
\icmlauthor{Jacob Andreas}{mit}
\icmlauthor{Shiyu Chang}{ucsb}
\icmlauthor{Yang Zhang}{ibm}
\end{icmlauthorlist}

\icmlaffiliation{ucsb}{UC Santa Barbara}
\icmlaffiliation{ucla}{UC Los Angeles}
\icmlaffiliation{mit}{MIT CSAIL}
\icmlaffiliation{ibm}{MIT-IBM Watson AI Lab}

\icmlcorrespondingauthor{Bairu Hou}{bairu@ucsb.edu}

\icmlkeywords{Machine Learning, ICML}

\vskip 0.3in
]



\printAffiliationsAndNotice{}  

\begin{abstract}
We describe \alg, a query-efficient procedure for building a text classifier from a neural language model (LM) without access to the LM's parameters, gradients, or hidden representations. 
This form of ``black-box'' classifier training has become increasingly important as the cost of training and inference in large-scale LMs has grown. But existing black-box LM classifier learning approaches are themselves computationally inefficient, typically specializing LMs to the target task by searching in a large space of (discrete or continuous) prompts using zeroth-order optimization methods.
Instead of directly optimizing in prompt space, {\alg} obtains a small pool of prompts via a gradient-free approach, and then constructs a large pool of \emph{weak learners} by pairing these prompts with different elements of the LM's output distribution. These weak learners are then ensembled using the \textsc{AdaBoost} algorithm. The entire learning process requires only a small number of forward passes per batch and no backward pass. Experiments show that {\alg} achieves state-of-the-art performance in multiple black-box few-shot classification tasks, and matches or outperforms full fine-tuning in both few-shot and standard learning paradigms,
while training 10x faster than existing black-box methods.Codes are available at \url{https://github.com/UCSB-NLP-Chang/PromptBoosting}.
\end{abstract}

\section{Introduction}

Prompt-based learning has emerged as an effective method to adapt pre-trained language models (LMs) for downstream natural language processing (NLP) tasks. A typical prompt-learning paradigm involves appending a specially-designed sequence, called a prompt, to the input to a pre-trained LM, which will thereby be repurposed for a given downstream task. Compared to standard fine-tuning, prompt-based learning is much more parameter-efficient.

Most prompt-based learning methods require searching for the optimal prompt for the downstream task. When gradient information of the pre-trained LM is available, such optimization can easily be performed by standard gradient-based methods~\citep{liu2021gpt, li2021prefix, lester2021power, zhang2021differentiable, liu2022ptuningv2}. However, in many real-world scenarios, the parameters, gradient, or hidden representations of the LMs are not accessible, \emph{a.k.a.} the black-box tuning setting, which makes gradient-based prompt learning very challenging~\citep{sun2022bbt}.

To tackle the challenges, the most common existing black-box solution is to resort to gradient-free optimization techniques to search for the optimal prompt, such as the zeroth-order gradient approximation~\citep{sun2022bbt, diao2022black} and reinforcement learning-guided optimization~\citep{deng2022rlprompt}. However, these methods would require a large number of queries to the LMs, which, considering the ever-growing size and computation cost of the pre-trained LMs, is highly inefficient and could lead to large approximation errors.
\begin{figure*}
    \centering
    \includegraphics[width=0.85\textwidth]{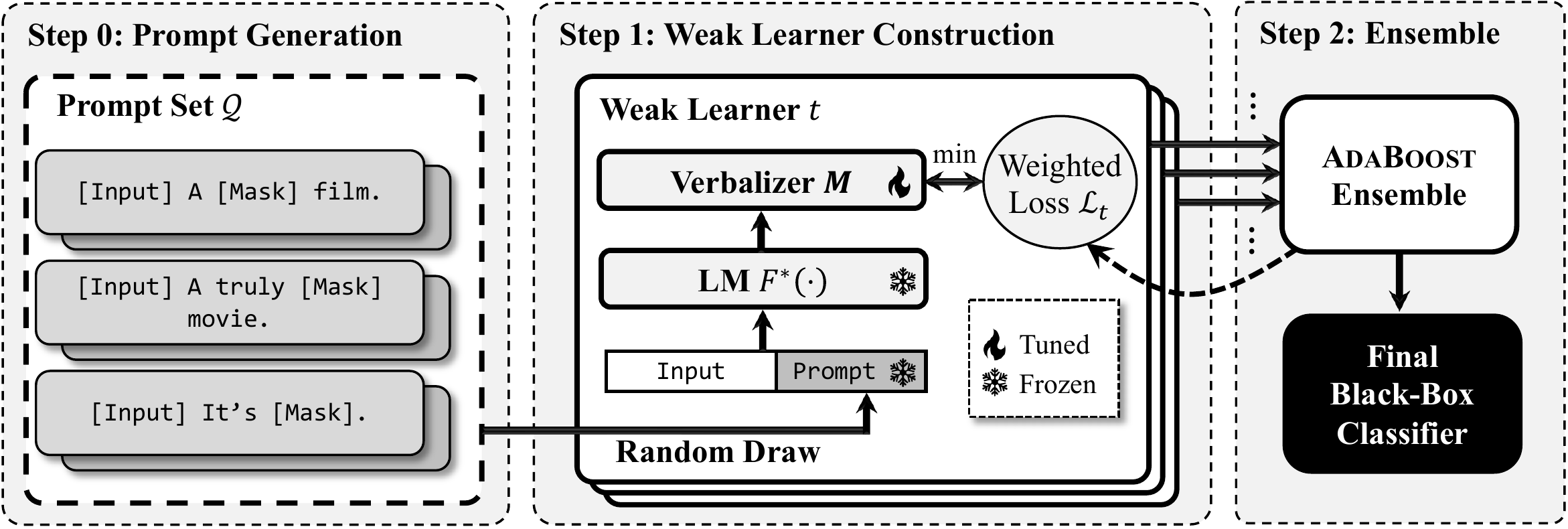}
    \vspace{-1mm}
    \caption{Overview of {\alg}.}
    \label{fig:overview}
\end{figure*}

In this paper, we propose \alg, a novel black-box prompt learning approach which does not rely on searching an optimal prompt and can thus drastically improve the computational efficiency over the existing methods. Figure\,\ref{fig:overview} illustrates the pipeline of {\alg}.  Specifically, rather than optimizing over the prompts, {\alg} constructs a small pool of prompts via a gradient-free approach. These prompts are sub-optimal because they are not optimized for any downstream tasks. Then, {\alg} creates a large pool of \emph{weak learners} by pairing each prompt with different elements of the LM's output distribution, which is commonly known as the \emph{verbalizer}. Finally, these weak learners are ensembled using the \textsc{AdaBoost} algorithm, where the optimization in each iteration is performed only over the verbalizer, not the prompt. The entire process only needs to evaluate the LM's output with each of the prompts, so it only involves a small number of forward passes per batch and no backward pass.

We evaluated our method on a number of downstream tasks. The results show that {\alg} achieves state-of-the-art performance and matches or even outperforms full fine-tuning in both few-shot and standard learning paradigms. Furthermore, {\alg} can run 10x faster than existing black-box prompt-learning approaches, with only ten forward passes per batch. 

\section{Related Work}
\textbf{Prompt-based learning} \ 
Prompt-based learning has emerged as a new approach for adapting pre-trained LMs for downstream tasks fueled by the success of GPT-3~\citep{brown2020language}. Since the prompts directly influence the performance of prompt-based learning, recent studies have focused on how to find the best prompts given a specific task. AutoPrompt~\citep{shin2020autoprompt} designs a gradient-based discrete optimization method to search for the optimal prompt. LM-BFF~\citep{gao2021making} leverages the pre-trained T5 model~\citep{raffel2020exploring} to automatically generate prompts and select the best one based on the performance on the validation set. Since verifying the automatically-generated prompts is time-consuming, the PTR~\citep{han2021ptr} method incorporates logic rules to construct prompts and to encode prior knowledge into prompt-based learning. 

Another line of work replaces the discrete prompt tokens with continuous embeddings that have their own parameters. P-tuning~\citep{liu2021gpt} trains a BiLSTM network to output continuous prompt embeddings. Prefix-tuning~\citep{li2021prefix} inserts prompt embeddings to each transformer layer in LMs and optimizes only the prompt embeddings during training. Prompt Tuning~\citep{lester2021power} also keeps the LMs frozen but adds prompt embeddings only in the input. P-tuning V2~\citep{liu2022ptuningv2} replaces the language model head in LMs with a linear layer for classification and shows that soft prompt tuning scales to medium-sized LMs and hard sequence tagging tasks. Our work adopts the discrete prompts for prompt-based learning.

\textbf{Black-box Tuning} \ 
Extremely large LMs such as GPT-3 are only provided as a service in the cloud, resulting inaccessible parameters and gradients of LMs. Furthermore, from the model provider's perspective, sharing hidden representations or gradients of LMs may reveal the vulnerability of the model and lead to security problems~\citep{tramer2016stealing}. How to find the optimal prompts in such a black-box tuning setting has attracted various explorations.
BBT~\citep{sun2022bbt} and BBTv2~\citep{sun2022bbtv2} employ the CMA evolution strategy, a derivative-free optimization method, to optimize continuous prompt embeddings. However, both of the two algorithms require querying the LM tens of thousands of times even in few-shot settings. 
Furthermore, both methods use soft prompts whereas the black-box setting typically only permits querying with textual input. Also, BBTv2 assumes that prompt embeddings can be added to each layer of the original language model, which is not accommodated in the standard black-box setting.
Clip-Tuning~\citep{chai2022clip} proposes to optimize the prompt embeddings on multiple subnetworks extracted from the original language model. While it outperforms BBT~\citep{sun2022bbt}, this method requires full access to the parameters of the language model.
RLPrompt~\citep{deng2022rlprompt} is a more realistic black-box tuning method where discrete prompt tokens are optimized through reinforcement learning and the performance on downstream tasks serves as the reward. 
BDPL~\citep{diao2022black} also utilizes the reinforcement learning method to optimize the discrete prompts but uses a lighter-weight policy network. It also narrows down the search space of prompt tokens by utilizing pointwise mutual information.
TEMPERA~\citep{zhang2022tempera} uses reinforcement learning to optimize discrete prompts and incorporates more components in optimization (\emph{e.g.}, exemplars for in-context learning).
GrIPS~\citep{prasad2022grips} performs phrase-level editing to generate discrete prompts. 
Existing black-box tuning methods suffer from poor efficiency and sub-optimal performance. Our method achieves high efficiency by first generating only a small set of prompts and achieves superior performance by then creating a set of weak learners from these prompts and ensembling them together via \textsc{AdaBoost}~\citep{freund1997decision}.

\textbf{Model ensemble} \
Model ensembling is a commonly used technique in machine learning. Prior to deep learning, Bagging~\citep{breiman1996bagging, breiman2001random} and Boosting~\citep{freund1997decision, friedman2001greedy} showed the power of model ensembling. One of these methods, \textsc{AdaBoost}~\citep{freund1997decision}, sequentially learns a series of weak learners and ensembles them for better generalization. During training, each weak learner is tweaked by leveraging examples that were misclassified by previous classifiers. Since the performance of each individual prompt can be weak, our method adopts \textsc{AdaBoost} as the framework for learning and ensembling multiple prompts.

\textbf{Prompt ensemble} \
As has been pointed out by prior work~\citep{lester2021power}, ensembling prompts is more efficient than ensembling entire fine-tuned models. 
Various ensemble strategies have been explored in past work. Uniformly averaging the predictions from different prompts has been used for factual probing~\citep{jiang2020can}, text generation~\citep{yuan2021bartscore, schick2020few}, and classification tasks~\citep{schick2021exploiting, lester2021power}. Furthermore, some methods adopt a weighted averaging strategy for better performance---the weight of each different prompt can be learned during training~\citep{jiang2020can, qin2021learning} or defined using some heuristics~\citep{schick2021exploiting, schick2021s}. Our method also falls into the prompt ensemble category. The main difference is each prompt-based model is sequentially learned conditioned on the classification errors of prior models.

\section{Methodology}

In this section, we will describe the {\alg} algorithm. For notation, we use $|\mathcal{A}|$ to denote the size of a finite set $\mathcal{A}$; $[A]$ to denote an index set $\{1, 2, \cdots, A\}$. 

\subsection{Problem Formulation}

Consider a text classification downstream task. Denote $\mathcal{D}_{\mathrm{tr}} = \bigcup_i \{(\bm x_i, y_i)\}$ as the training set, where $\bm x_i$ denotes the input text sequence and $y_i$ denotes the output label. We are given a pre-trained language model, denoted as $\bm p_i = F^*(\bm x_i)$, which, given the input $\bm x_i$, produces a probability distribution over the vocabulary set, $\mathcal{V}$, at a given location. In this paper, the output distribution is relevant only where the input is \texttt{[mask]}, so $\bm p_i \in \mathbb{R}^{|\mathcal{V}| \times 1}$ is just a $|\mathcal{V}|$-dimensional vector specifying the output probability at the \texttt{[mask]} location, where $|\mathcal{V}|$ denotes the vocabulary size. Our goal is to adapt the LM $F^*(\cdot)$ to the downstream task using the downstream training set $\mathcal{D}_{\mathrm{tr}}$.

We adopt the common prompt-learning framework, where the parameters of $F^*(\cdot)$ are frozen (we add a superscript $^*$ to emphasize this). The following two mechanisms are added to convert $F^*(\cdot)$ into a text classifier for the given downstream tasks.
\vspace*{-2mm}
\begin{enumerate}[leftmargin=*]
    \item \textbf{Prompt} \quad A prompt is a sequence of tokens that is concatenated to the input. Formally, denote the prompt sequence as $\bm q$ and the concatenated input sequence as $\bm x_i \Vert \bm q$. Then the LM is modified as $F^*(\bm x_i \Vert \bm q)$.
\vspace*{-1mm}
    \item \textbf{Verbalizer} \quad To convert the output probability over the \emph{vocabulary} into that over the \emph{classes}, a verbalizer is introduced to assign each token into different classes. Formally, denote the number of classes of the downstream task as $|\mathcal{Y}|$, then the verbalizer is a $|\mathcal{Y}|$-by-$|\mathcal{V}|$ matrix, denoted as $\bm M$, where the element in row $c$, column $v$ represents the assignment weight of the $v$-token in the vocabulary into class $c$. Each row of $\bm M$ would sum up to one. The predicted probability of all the classes can then be expressed as $\bm M \bm p_i$.
\end{enumerate}
\vspace*{-2mm}

To sum up, after the prompt and verbalizer are applied, the adapted LM becomes $\bm M F^*(\bm x_i \Vert \bm q)$. Therefore, the prompt-tuning process boils down to learning an appropriate verbalizer $\bm M$ and prompt $\bm q$ for the downstream task. 

\subsection{Algorithm Overview}
\label{subsec:overview}

Conventional black-box prompt learning methods commonly use a pre-set $\bm M$ while performing black-box optimization over $\bm q$, which results in a large computation cost. In contrast, {\alg} randomly chooses from a small number of pre-generated prompts and performs optimization over $\bm M$ instead. Due to the sub-optimality of pre-generated prompts and the limited representation power of $\bm M$, the resulting classifiers are weak. However, this process is able to quickly generate a large pool of such weak learners, which can then be ensembled into a strong learner using the \textsc{AdaBoost} approach. As the optimization over $\bm M$ is computationally cheap, the ensemble process is much more efficient than the conventional black-box methods.

More specifically, {\alg} iteratively generates $T$ weak learners, and each weak learner $t$ is optimized under its respective loss function, denoted as $\mathcal{L}_t(\bm q, \bm M)$, which is essentially a weighted loss over the training set with larger weights on those that are misclassified by the previous weak learners (More details of the \textsc{AdaBoost} algorithm will be provided in Section~\ref{subsec:ensemble}). As shown in Figure~\ref{fig:overview}, {\alg} consists of the following key steps.

\textbf{Step 0:} Generate a pool of prompts, $\mathcal{Q} = \bigcup_j \{\bm q_j\}$, using a gradient-free method.

\textbf{Step 1:} Construct $T$ weak learners. For weak learner $t$, its prompt $\bm q_t$ is uniformly randomly drawn from $\mathcal{Q}$; its verbalizer $\bm M$ is determined by solving 
\begin{equation}
\small
\begin{aligned}
     \min_{\bm M} \mathcal{L}_t(\bm q_t, \bm M), \ \ \ \
     & \mbox{s.t.} \quad \bm M_{cv} \geq 0, \forall c \in [|\mathcal{Y}|], v \in [|\mathcal{V}|],\\
     & \ \ \ \ \ \ \sum_{c \in [|\mathcal{Y}|]} \bm M_{cv} = 1, \forall v \in [|\mathcal{V}|]
\end{aligned}
\label{eq:obj}
\end{equation}

\textbf{Step 2:} Ensemble the weak learners using \textsc{AdaBoost}.

Section~\ref{sec: verbalizer_learning} will describe how to solve \eqref{eq:obj}. Section~\ref{subsec:prompt_set} will describe how the pool of prompts, $\mathcal{Q}$, is generated.

\subsection{Learning the Verbalizer}
\label{sec: verbalizer_learning}
As discussed, the loss function $\mathcal{L}_t$ as in \eqref{eq:obj} is essentially a weighted sum of the individual loss over the training dataset $\mathcal{D}_{\mathrm{tr}}$, \emph{i.e.}
\begin{equation}
\small
    \mathcal{L}_t(\bm q_t, \bm M) = \sum_{(\bm x_i, y_i) \in \mathcal{D}_{\mathrm{tr}}} w_{ti} \ell(\bm x_i, y_i; \bm q_t, \bm M),
    \label{eq:weighted_loss}
\end{equation}
where $w_{ti}$ denotes the weight on training data point $i$ for learning weak learner $t$ as determined by \textsc{AdaBoost}; $\ell(\bm x_i, y_i; \bm q_t, \bm M)$ denotes the loss on data point $(\bm x_i, y_i)$ with the parameters set to $\bm q_t$ and $\bm M$. Since we focus on classification tasks, $\ell(\cdot)$ should ideally be the cross-entropy loss. However, the optimization problem in \eqref{eq:obj} is essentially a partition problem, which can easily lead to combinatorial complexity. To derive the tractable solution, we adopt the following strategy. First, solve \eqref{eq:obj} with $\ell(\cdot)$ set to the $\ell_1$ loss, which, though not optimal for the classification task, bears a closed-form solution. Second, further screen the token assignment by maximizing the training set performance. The detailed method is described below.

\paragraph{Minimizing the $\ell_1$ loss} 
By replacing the $\ell(\cdot)$ in \eqref{eq:weighted_loss} with the $\ell_1$ loss, a closed-form solution can be derived, which can establish a basis for the subsequent steps for deriving a good verbalizer. Formally, let $\bm h_i$ be the one-hot representation of the class label $y_i$, and let $\bm \pi_i =  F^*(\bm x_i \Vert \bm q_t)$ represent the LM output probability with the prompt $\bm q_t$ concatenated. Then, 
with the $\ell_1$ loss, \eqref{eq:weighted_loss} becomes
\begin{equation}
\small
\begin{aligned}
    \mathcal{L}_t(\bm q_t, \bm M) &= \sum_{(\bm x_i, y_i) \in \mathcal{D}_{\mathrm{tr}}} w_{ti} \lVert\bm h_i - \bm M \bm \pi_i \rVert_1 \\
    &= \sum_{(\bm x_i, y_i) \in \mathcal{D}_{\mathrm{tr}}} w_{ti} \bm{1}^T |\bm h_i - \bm M \bm \pi_i | \\
    &= \sum_{(\bm x_i, y_i) \in \mathcal{D}_{\mathrm{tr}}} w_{ti} \big[(-\bm{1})^{\bm h_i}\big]^T (\bm M \bm \pi_i - \bm h_i).
\end{aligned}
\label{eq:l1_loss}
\end{equation}
Here, $\bm{1}$ represents a all-one column vector of dimension $|\mathcal{Y}|$, and $(-\bm{1})^{\bm h_i}$ represents the element-wise power operation. The last equality is because each element of $\bm M \bm \pi_i$ is within $[0, 1]$ and each element of $\bm h_i$ is either $0$ or $1$, 
so we can easily remove the absolute sign depending on the actual values of $\bm h_i$.

As shown in \eqref{eq:l1_loss}, the loss function is \emph{linear} with respect to $\bm M$, so the optimization in \eqref{eq:obj} becomes a linear optimization problem with linear constraints, which has closed-form corner solutions. For notational brevity, define a score matrix, $\bm S$, as
\begin{equation}
    \small
    \bm S = \sum_{(\bm x_i, y_i) \in \mathcal{D}_{\mathrm{tr}}} w_{ti} \bm \pi_i \big[(-\bm{1})^{\bm h_i}\big]^T,
\end{equation}
which is the same size as $\bm M$ and is essentially the coefficients multiplied with $\bm M$ in \eqref{eq:l1_loss}. Then, we state without detailed derivations that the solution to \eqref{eq:obj} is such that each token is assigned to the class for which it gets the highest score among all the classes, \emph{i.e.,}
\begin{equation}
    \bm M_{cv} = 1, \quad \mbox{if } c = \argmax_{c'\in[|\mathcal{Y}|]} \bm S_{c'v}, \quad \mbox{and } 0 \mbox{ otherwise}.
    \label{eq:l1_solution}
\end{equation}
Since the $\ell_1$ loss does not generally work well for classification tasks, we empirically find that the verbalizer derived in \eqref{eq:l1_solution} is of limited performance. However, this inspires us that the score matrix, $\bm S$, is a good measure of how well each token should be selected for a class. In the following step, we will further screen the tokens with the help of the score matrix.

\paragraph{Screening the tokens} 
One issue with the verbalizer in \eqref{eq:l1_solution} is that each token has to be assigned to one class, even those tokens that are not good indicators of any particular class. Therefore, by removing the non-informative tokens and only retaining the best tokens for each class, we can improve the verbalizer performance. To reduce the computational complexity, we will retain only one token for each class. Specifically, we first identify a candidate set of tokens for each class by choosing the tokens with top-$m$ scores for that class, \emph{i.e.,} the top-$m$ elements in $\bm S_{c:}$ for class $c$, where subscript $c\colon$ denotes the $c$-th row. Then, we evaluate all the possible combinations that include one token from the candidate set for each class (hence $m^{|\mathcal{Y}}|$ combinations in total) and choose the combination that achieves the best training accuracy (weighted by $\{w_{ti}\}$).

\subsection{Constructing the Prompt Set}
\label{subsec:prompt_set}
To generate the pool of prompts, $\mathcal{Q}$ (step 0 in section~\ref{subsec:overview}), we adopt the optimization-free method proposed by \citet{gao2021making}, which employs the T5~\citep{raffel2020exploring} model. Specifically, we first construct a small subset of the training set, denoted as $\mathcal{D}_{\mathrm{gen}}$, to induce the prompt generation ($\mathcal{D}_{\mathrm{gen}}$ is exactly $\mathcal{D}_{\mathrm{tr}}$ in few-shot setting). Then, for each data point $(\bm x_{i},y_{i}) \in\mathcal{D}_{\mathrm{gen}}$, we construct an input to the T5 model as 
{\small\texttt{<input><A><label token><B>}}
(for sentence-pair classification tasks, the input to T5 becomes
{\small\texttt{<input1><A><label token><B><input2>}}).
Here, \texttt{\small <A>} and \texttt{\small<B>} are mask tokens in T5 representing spans to be filled in.  \texttt{\small<input>}, \texttt{\small<input1>} and \texttt{\small<input2>} represent the input text $\bm x_i$.
\texttt{\small<label token>} is a pre-defined mapping to convert class labels to tokens in $\mathcal{V}$. For example, positive label ($y_i = 1$) in SST-2~\citep{socher2013recursive} dataset is mapped to token \texttt{\small great} while negative label ($y_i = 0$) is mapped to \texttt{\small terrible}. Given this input, the T5 model fills in the spans for \texttt{\small<A>} and \texttt{\small<B>}. The decoding process aims to maximize output probability conditioned on the input over $\mathcal{D}_{\mathrm{gen}}$. 
Then the T5 generated outputs, denoted as \texttt{\small<output A>} and \texttt{\small<output B>} will be converted into prompts and concatenated to the training input text, \emph{i.e.,} $\bm x_{i} \Vert \bm q$, in the form of 
\texttt{\small<input><output A>[mask]<output B>}
(for sentence-pair tasks, the form becomes \texttt{\small<input1><output A>[mask]<output B><input2>}).
As an example, on SST-2 dataset, one of the generated outputs by T5 is \texttt{\small<output A>} = \texttt{\small A truly}, \texttt{\small<output B>} = \texttt{\small movie}. Then the input sentence ``\texttt{\small I love it.}'' will be converted to ``\texttt{\small I love it. A truly [MASK] movie}''.
With a wide beam search width (by default we use $100$), we select the top-10 generated prompts according to the log-likelihood to form the prompt pool, $\mathcal{Q}$. All the generated prompts used in our experiments can be found in Table\,\ref{tab: template_visualize} in Appendix\,\ref{sec: templates_visualize}. Readers can refer to \citet{gao2021making} for further details. The entire generation process does not involve any optimization over the prompts, and thus is computationally efficient. 
It is worth noting that the aforementioned approach can be replaced with any other optimization-free prompt generation methods, such as manually creating the prompts, making {\alg} flexible for realistic use. 
Our experiment results in Table\,\ref{tab:prompt_study} in Section\,\ref{subsec: eval_result} show that our method works well with different prompt generation methods.
\subsection{Ensembling the Weak Learners}
\label{subsec:ensemble}
We follow the \textsc{AdaBoost} algorithm to ensemble the weak learners. As discussed, each weak learner minimizes a weighted loss over the training set (\eqref{eq:weighted_loss}). The final prediction is produced by taking a weighted average over the weak classifiers' output. Further details, including how the weights are computed, are shown in Algorithm \ref{alg: adaboost}. 
\begin{minipage}{\linewidth}
\vspace{-1mm}
\begin{algorithm}[H]
\small
\caption{\small Model Ensemble in {\alg}}
\label{alg: adaboost}
\begin{algorithmic}[1]
\STATE \textbf{Input:} prompt set $\mathcal{Q}=\bigcup_j \{\bm q_j\}$, LM $F^{*}(\cdot)$, $\mathcal{D}_{\mathrm{tr}}$,
\STATE \textbf{Output:} weak learners $\bigcup_{t} \{f_{t}(\cdot)\}$ and their weights $\bigcup_{t} \{\alpha_{t}\}$.
\STATE Set initial data weight to $w_{1i} = 1 / |\mathcal{D}_{\mathrm{tr}}|$, $\forall i \in [|\mathcal{D}_{\mathrm{tr}}|]$
  \FOR{Iteration $t=1,\ldots,T$}
  \STATE Randomly draw a prompt $\bm q_{t}$ from $\mathcal{Q}$
  \STATE Learn the verbalizer $\bm M_t$ with weight $\{w_{ti}\}$
  \STATE Set weak learner $t$ to $f_t(\cdot) = \bm M_t F^*(\cdot \Vert \bm q_t)$
  \STATE Compute weighted error as

  $err^{(t)} = \sum_{i=1}^{|\mathcal{D}_{\mathrm{tr}}|}w_{ti}\1_{\mathrm y_{i}\ne f_t(x_{i})}/\sum_{i=1}^{|\mathcal{D}_{\mathrm{tr}}|}w_{ti}$
  \STATE Compute the weight on $f_t$ as
  
  $\alpha_{t} = \log \frac{1-err^{(t)}}{err^{(t)}} + \log (|\mathcal{Y}|-1)$
  \STATE Adjust dataset weight
  
  $w_{(t+1)i} = w_{ti}\cdot \exp(\alpha_{t}\cdot \1_{y_{i}\ne f_{t}(x_{i})})$, $\forall i \in [|\mathcal{D}_{\mathrm{tr}}|]$
  
  \STATE Re-normalize $\{w_{(t+1)i}\}$.
  
  \ENDFOR
\end{algorithmic}
\end{algorithm}
\end{minipage}

It is worth mentioning that we can generate many weak learners at a very low computational cost because we only need to evaluate the LM's output distribution with each of the pre-generated prompts in $\mathcal{Q}$, beyond which no extra forward pass is needed when learning each weak learner. Since the number of pre-generated prompts is small, typically ten in our implementation, the entire learning process involves no more than ten forward passes per batch in the training set, no matter how many weak learners are generated.

\section{Experiments}
\subsection{Experiment Setup}
\label{subsec: setup}
\noindent \textbf{Datasets} \ 
Previous approaches for black-box prompt-based learning~\citep{sun2022bbt,sun2022bbtv2,deng2022rlprompt, zhang2022tempera} are often evaluated on the following tasks: single sentence classification (including SST-2 \citep{socher2013recursive}, MR \citep{pang2005seeing}, TREC \citep{voorhees2000building} and AG's News \citep{Zhang2015CharacterlevelCN}) and sentence-pair classification (including SNLI \citep{bowman2015large}, MNLI-m \citep{williams2018broad}, QNLI \citep{rajpurkar2016squad}, and RTE \citep{dagan2005pascal}). We follow the same setting and report results on the datasets above. The dataset statistics can be found in Table\,\ref{tab: dataset_stat} in Appendix \ref{sec: implementation}. 
For a more comprehensive understanding of our method, we incorporate additional datasets including SST-5~\citep{socher2013recursive},
CR~\citep{hu2004mining}, Subj~\citep{pang2004sentimental}, MPQA~\citep{wiebe2005annotating}, MRPC~\citep{dolan2005automatically} in Table\,\ref{tab:all_dataset} in Appendix\,\ref{sec: addtion_exp}---the conclusion is the same.

\noindent \textbf{Evaluation setting} \ 
We mainly evaluate the performance of {\alg} in few-shot settings. This is reasonable especially for black-box model tuning scenarios, where the maximum allowed query times may be limited.
We randomly sample $k$ examples per class from the original training set to construct a $k$-shot training set $\mathcal{D}_{\mathrm{tr}}$ for model training. 
Following previous work~\citep{gao2021making,zhang2021differentiable, sun2022bbt}, we also construct the validation set $\mathcal{D}_{\mathrm{val}}$ by randomly sampling another $k$ examples per class from the original training set (i.e., $|\mathcal{D}_{\mathrm{tr}}| = |\mathcal{D}_{\mathrm{val}}|$). By default we set $k = 16$ for our main experiments. 
Also, while previous work splits the training and validation sets in this way and we do so for direct comparison, we also explore integrating the validation set into the training set—in a truly few-shot setting, we should make full use of as many examples as we can, and we show this leads to an improvement in performance.
As for evaluation, we use the whole testing set. For SNLI~\citep{bowman2015large} and the datasets from the GLUE benchmark~\citep{wang2018glue}, we use the original validation set for evaluation.

\noindent \textbf{Backbone models} \ 
In the main experiments, we adopt the widely-used RoBERTa-large model~\citep{liu2019roberta} for evaluation to allow for direct comparison with baselines.

\noindent \textbf{Baselines} \ 
We compare {\alg} with fine-tuning and state-of-the-art black-box tuning methods described below. For reference, we also include white-box prompt-based learning methods that are designed for a few-shot setting. Implementation details can be found in Appendix \ref{sec: implementation}. 
\textbf{(1) Fine-tuning} is just standard model fine-tuning in a few-shot setting.
\textbf{(2) LM-BFF}~\citep{gao2021making} is a prompt-based fine-tuning method. In LM-BFF, all input will be transformed using automatically generated prompts. Then the whole model is fine-tuned based on the transformed data.
\textbf{(3) DART}~\citep{zhang2021differentiable} replaces the discrete prompts in LM-BFF with trainable prompt embeddings, which can reduce the prompt generation cost.
\textbf{(4) BBT}~\citep{sun2022bbt} employ zeroth-order gradients to optimize the continuous prompts.
\textbf{(5) BBTv2}~\citep{sun2022bbtv2} improves the performance of BBT by inserting prompt embeddings into each layer of the language model
\textbf{(6) RLPrompt}~\citep{deng2022rlprompt} models the black-box optimization of discrete prompts as a reinforcement learning problem and adopts Q-learning to find the best prompt.
Some black-box baselines~\citep{zhang2022tempera, chai2022clip} are not included because the official implementation is not available.
\begin{table*}[t]
    \centering
    \caption{\footnotesize{Performance of proposed {\alg} and baseline methods in few-shot setting ($k=16$) measured by classification accuracy (\%). All methods use RoBERTa-large~\citep{liu2019roberta} as the backbone LM for a fair comparison. 
    Two white-box methods are included for reference including LM-BFF and DART. BBT, BBTv2, and RLPrompt are the main black-box baselines. {\alg}-32 combines both training and validation sets for training.
    Mean accuracy (and standard deviation) is reported over 5 different splits. The best results are highlighted in \textbf{bold} and the second best are \underline{underlined}.}}
    \label{tab:fewshot_exp}
    \vspace{1mm}
    \resizebox{0.95\textwidth}{!}{
    \begin{tabular}{l|ccccccccc}
    \toprule[1pt]
         Method & SST-2 & MR & AG's News & TREC & SNLI & MNLI & QNLI & RTE & Avg.  \\
    \midrule
         Fine-tuning & 81.4 (3.8)& 82.7 (3.6)& 86.2 (1.4) & 88.8 (2.1) & 48.4 (4.8) & 45.8 (6.4) & 56.3 (1.5) & 54.4 (3.9) &  68.0\\
         LM-BFF~\citep{gao2021making} & 92.3 (1.5)& 87.4 (0.6) & 87.1 (1.2) & 83.4 (2.7) & 76.5 (2.6)& 68.7 (2.0) & 64.4 (4.6) & 66.6 (6.4)& 78.3 \\
         DART~\citep{zhang2021differentiable} & 93.5 (0.5) & 88.2 (1.0) & 86.8 (0.5) & 87.1 (3.8)& 75.8 (1.6) & 67.5 (2.6) & 66.7 (3.7) & 59.0 (2.5)&  78.1\\
    \midrule
         BBT~\citep{sun2022bbt} & 88.2 (1.7) & 82.8 (2.6) & 81.2 (2.7) & 39.3 (5.2)& 44.7 (4.0)& 42.3 (2.8) & 56.8 (2.0) & 49.1 (3.3) & 60.6 \\
         BBTv2~\citep{sun2022bbtv2} & \underline{88.5 (2.1)} & 83.7 (1.8) & 83.6 (2.0) & 63.8 (9.9) & 57.4 (2.7) & 51.4 (3.3) & \underline{58.1 (2.5)} & 53.2 (7.0) & 67.5 \\
         RLPrompt~\citep{deng2022rlprompt} & \textbf{90.5 (1.5)} & \textbf{86.2 (2.5)} & 76.2 (2.7)& 37.3 (3.5)& 42.9 (1.8) & 40.7 (4.7)& 52.1 (2.9) & 52.2 (2.2) &  59.8\\
\rowcolor{Gray}
         {\alg} & 87.6 (3.0) & \underline{84.6 (2.5)} & \textbf{85.2 (0.9)} & \underline{81.6 (4.0)} & \underline{61.3 (3.5)}& \underline{52.5 (1.5)} & 58.0 (3.3) & \underline{60.0 (5.5)} &  \underline{71.4}\\
\rowcolor{Gray}
         {\alg}-32 & 87.6 (3.3) & 84.7 (2.1) & \underline{84.2 (1.1)} & \textbf{84.5 (1.4)} &  \textbf{62.0 (2.7)} & \textbf{53.8 (1.2)} & \textbf{58.3 (2.8)} & \textbf{60.3 (2.4)} &  \textbf{71.9}\\
    \bottomrule[1pt]
    \end{tabular}
    }
\vspace{-3mm}
\end{table*}

\begin{table*}[t]
    \centering
    \caption{\footnotesize{Deployment efficiency of proposed {\alg} and baseline methods in few-shot setting ($k=16$). With all methods using RoBERTa-large (335M parameters) as the backbone LM, some baselines introduce additional parameters, leading to a slight variation in total parameters. Wall time is reported to measure the training time efficiency. Query efficiency is evaluated by \#Forward and \#Backward, which refer to the number of forward/backward passes per batch during training respectively.}}
    \vspace{1mm}
    \label{tab:efficiency}
    \resizebox{0.95\textwidth}{!}{
    \begin{tabular}{l|cc|cccc|cccc}
    \toprule[1pt]
         \multirow{2}{*}{Method} & \multirow{2}{*}{\makecell[c]{Trainable\\param}} & \multirow{2}{*}{\makecell[c]{Total\\param}} & \multicolumn{4}{c|}{AG's News} & \multicolumn{4}{c}{RTE}  \\
         & & & Acc & Wall Time & \#Forward &\#Backward & Acc & Wall Time & \#Forward &\#Backward \\
    \midrule
         Fine-tuning & 335M & 335M & 86.2 & 13 min & 100 & 100 & 54.4 & 19 min & 100 & 100 \\
         LM-BFF~\citep{gao2021making} & 335M & 335M & 87.1 & 5 min & 32 & 32 & 66.6 & 9 min & 60 & 60\\
         DART~\citep{zhang2021differentiable} & 335M & 335M & 86.8 & 15 min & 30 & 30 & 59.0 & 5 min & 120 & 120 \\
    \midrule
         BBT~\citep{sun2022bbt} & 25k & 335M & 81.2 & 88 min& 8,000 & 0 & 49.1 & 52 min & 8,000 & 0 \\
         BBTv2~\citep{sun2022bbtv2} & 25k & 335M & 83.6 & 90 min & 8,000 & 0 & 53.2 & 70 min & 8,000 & 0 \\
         RLPrompt~\citep{deng2022rlprompt} & 3M & 420M & 77.2 & 117 min & 1,000 & 0 & 52.2 & 90 min & 1,000 & 0 \\
\rowcolor{Gray}
         {\alg} & $<$1k & 335M & 85.2 & 8 min & 10 & 0 & 60.0 & 4 min & 10 & 0\\

    \bottomrule[1pt]
    \end{tabular}
    }
\vspace{-3mm}
\end{table*}

\noindent \textbf{Implementation details} \ 
We use the official implementations and hyper-parameters for all baselines.
For more details, please refer to Appendix\,\ref{sec: implementation}. For our method, we sequentially train 200 weak classifiers on each task and add them to our ensemble---we stop when validation performance plateaus or when we reach the maximum number of weak classifiers.

\subsection{Evaluation Results}
\label{subsec: eval_result}
\noindent \textbf{Overall comparison} \ 
We first evaluate the effectiveness of {\alg} in a few-shot setting with experiment results in Table\,\ref{tab:fewshot_exp}.
Although there is some variance across datasets, {\alg} achieves state-of-the-art performance compared to existing black-box tuning methods.

We emphasize the effectiveness of model ensembling in {\alg}. Firstly, on the SST-2 and MR datasets, which are sentiment analysis tasks, even individual weak learners in {\alg} can achieve 100\% accuracy on the training set, making the model ensemble inapplicable (note that AdaBoost cannot ensemble classifiers that achieve 100\% accuracy). Therefore, we directly train 10 weak learners using 10 prompts on the unweighted training set and then select the weak learner that performs best on the validation set as the final model. Since the advantage of model ensemble is limited on SST-2 and MR datasets, it is not surprising that {\alg} performs slightly worse than BBT and RLPrompt. However, {\alg} is still better than fine-tuning an MLP, demonstrating the effectiveness of our proposed verbalizer learning method. 

Secondly, on the other 6 datasets, {\alg} consistently outperforms all the baselines, with only one exception in the QNLI dataset, where BBTv2 has a slight advantage. However, notice that BBTv2 has an unfair advantage over all the other black-box methods, including ours, in allowing soft prompts to be added to each intermediate layer of the language model.
{\alg} also outperforms standard fine-tuning on the 4 NLI tasks. It is worth noting that on the TREC dataset, all of the black-box baselines performs very badly except for {\alg}, which even achieves a level of accuracy close to that of white-box methods. One potential reason is that the TREC dataset is harder for prompt-based learning. For example, the manual prompt on the TREC dataset achieves only 32\% accuracy~\cite{gao2021making}. According to our experiments, individual weak learners trained on the unweighted training set using our verbalizer learning method can only achieve 30\%-50\% accuracy. However, after model ensembling, the performance is largely improved, demonstrating the effectiveness of {\alg}.

Finally, we incorporate a variant of {\alg}, namely {\alg}-32, which skips the hyper-parameter tuning and directly integrates the validation set into training. The hyper-parameter, i.e., the number of weak classifiers, is determined manually according to its value when the validation set is available. Expanding the training set gives a slight improvement in the performance and decreases the variance.

\noindent \textbf{Deployment efficiency} \ 
Another concern with black-box model tuning is the deployment efficiency. As we have discussed above, directly adopting zeroth-order gradient optimization techniques suffers from the need to query many times, making it less applicable in realistic scenarios. We visualize the deployment efficiency of different methods in Table\,\ref{tab:efficiency}. AG's News and RTE datasets are adopted due to the average input length (see Table \ref{tab: dataset_stat}). The metrics include parameter efficiency (number of trainable parameters and total parameters), wall time of training, and number of forward/backward passes per batch. In terms of trainable parameters, {\alg} optimizes only less than 1k parameters ($|\mathcal{Y}| * 200$) and does not introduce any extra parameters. In contrast, RLPrompt uses another network, DistilGPT2~\citep{sanh2019distilbert}, in addition to the backbone RoBERTa model and consequently increases the training cost. In terms of wall time, {\alg} improves the efficiency over existing black-box tuning baselines (BBT, BBTv2, and RLPrompt) by more than 10 times. The query time is also significantly lower. Only 10 forward passes per batch of training data are required during the training of {\alg}. By contrast, our baselines require thousands of forward passes, which makes them hard to use in realistic scenarios. 
In addition, we can further significantly improve the efficiency of {\alg} with some slight simplifications without hurting the performance. Please refer to Table\,\ref{tab:improve_efficiency} in Appendix\,\ref{sec: addtion_exp} for more details.
\begin{figure*}[t]
    \centering
    \subfigure[\footnotesize{Performance on SST-2}]
    {          
        \includegraphics[width=.23\textwidth, height=!]{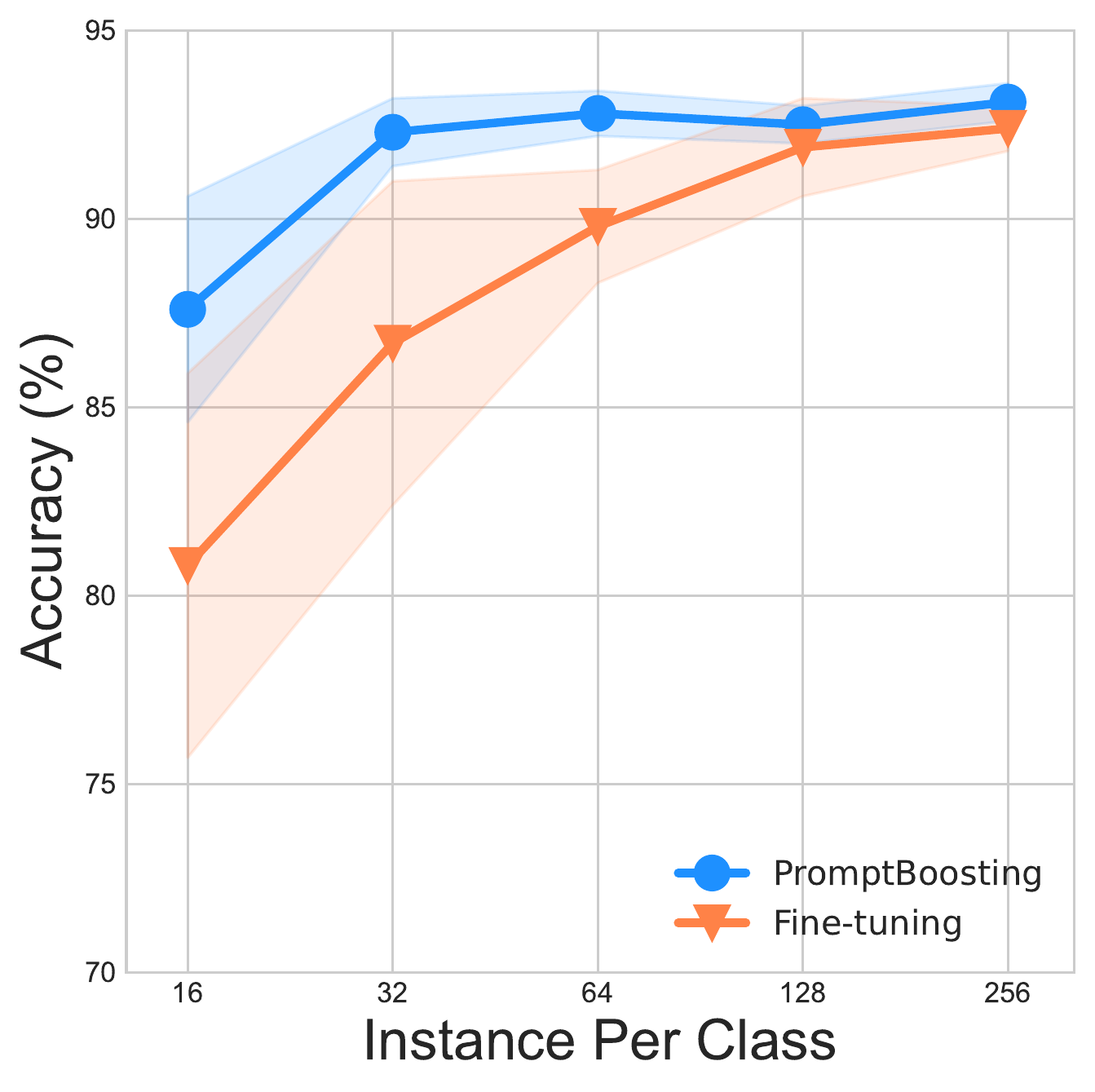}
    }
    \hspace{-1mm}
    \subfigure[\footnotesize{Performance on MR}]
    {          
        \includegraphics[width=.23\textwidth, height=!]{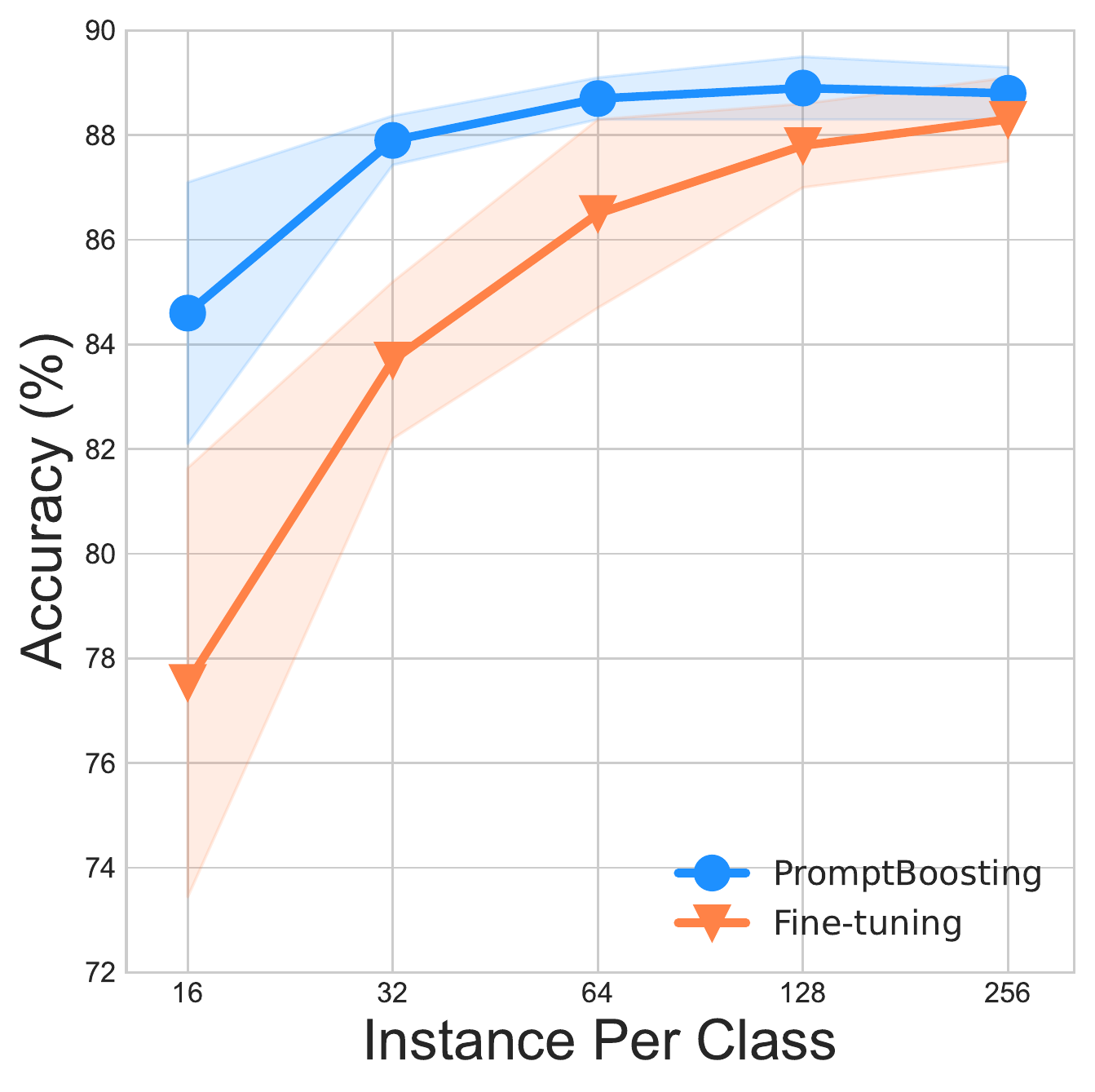}
    }
    \hspace{-1mm}
    \subfigure[\footnotesize{Performance on SNLI}]
    {          
        \includegraphics[width=.23\textwidth, height=!]{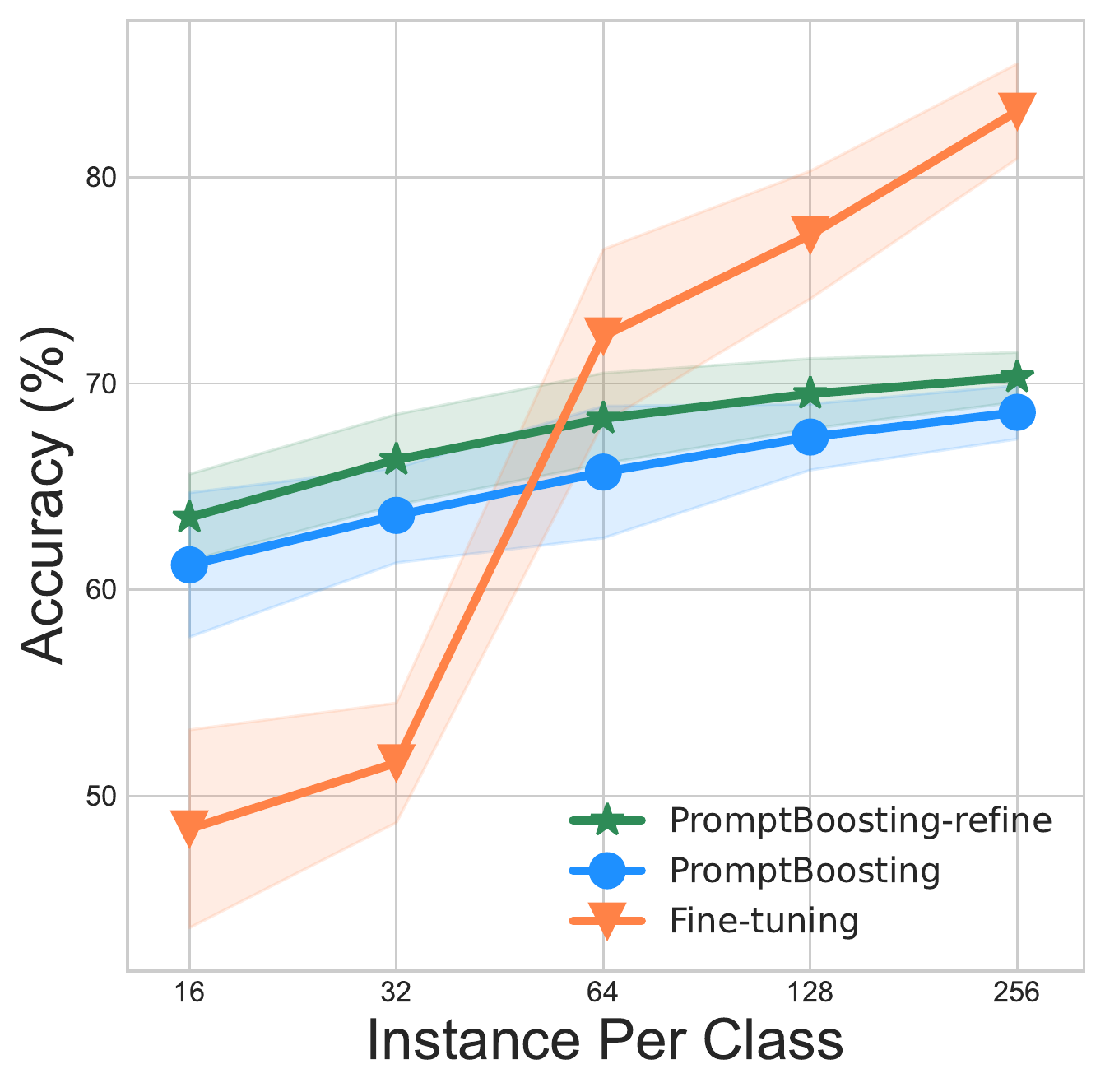}
    }
    \hspace{-1mm}
    \subfigure[\footnotesize{Performance on MNLI}]
    {          
        \includegraphics[width=.23\textwidth, height=!]{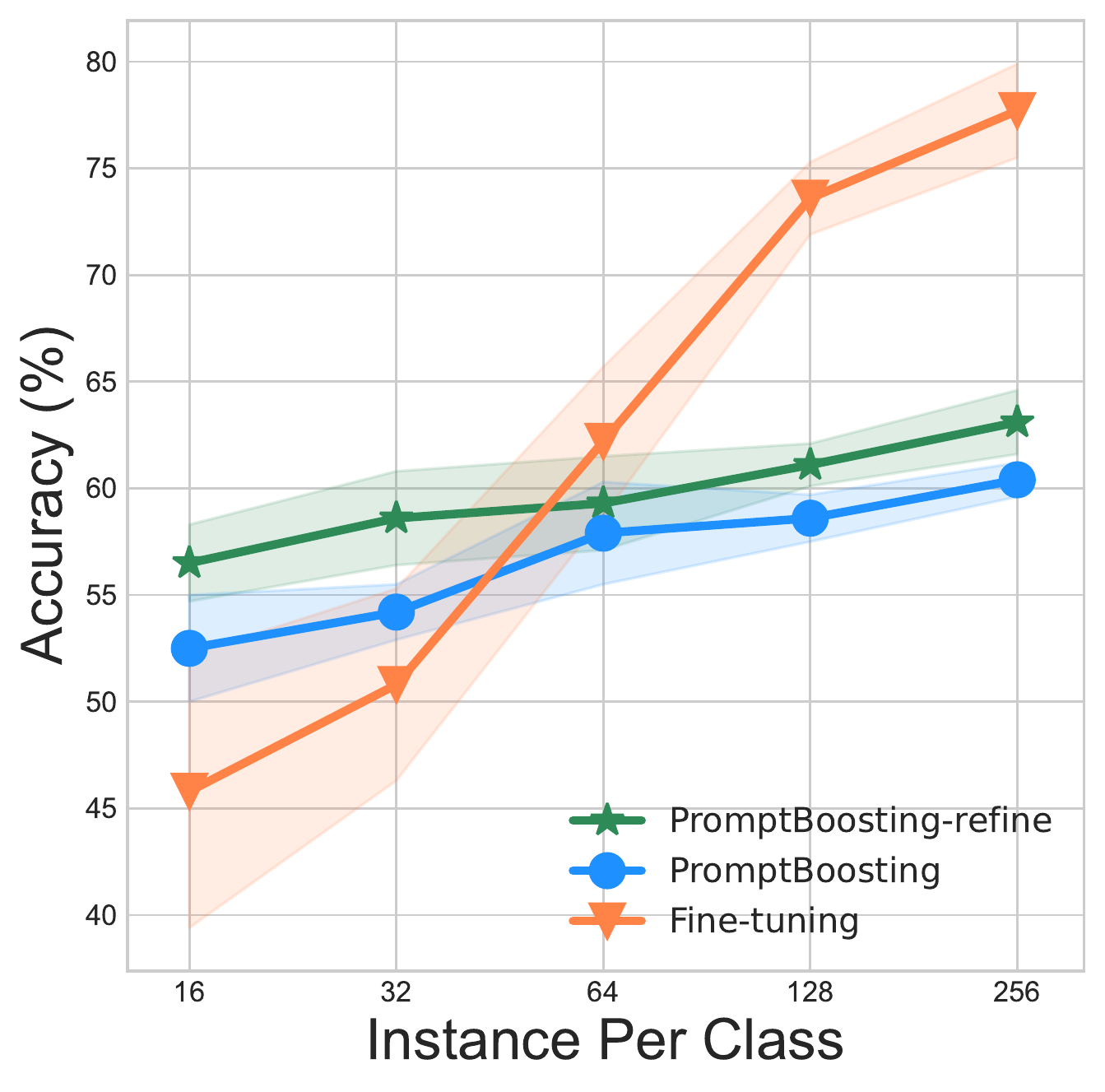}
    }
\vspace{-2mm}
    \caption{Model performance as a function of training set size on different datasets. For NLI tasks (SNLI and MNLI), we also include prompt refinement for better performance.}
\label{fig:datasize}
\vspace{-4mm}
\end{figure*}

\noindent \textbf{Effect of training data size} \ 
We also study the performance of {\alg} as the size of the training set increases (see Figure\,\ref{fig:datasize}). Note that we still fix $k = 16$ for the validation set regardless of the training set size. Results on AG's News, TREC, QNLI, and RTE dataset are shown in Figure\,\ref{fig: appendix_training_size} in Appendix\,\ref{sec: addtion_exp}. The conclusions are in three dimensions. Firstly, on the SST-2 and MR datasets, {\alg} consistently outperforms fine-tuning with lower variance, demonstrating the effectiveness of our method. Secondly, on the AG News and TREC datasets, {\alg} performs worse than fine-tuning. A similar phenomenon also exists in past work~\citep{gao2021making}, where even a white-box prompt-based few-shot learning method can achieve performance that is at most only comparable with fine-tuning. However, we remark that our method still maintains large advantages compared to all black-box baseline methods and achieves highly usable performance. Finally, as the amount of training data increases, the performance of fine-tuning improves and gradually outperforms our method on the four NLI datasets. This finding is possibly due to the fact that pre-trained LMs before fine-tuning are not good at tasks involving sentence pairs.

\noindent \textbf{Refinement of prompts} \ 
The performance of the weak learner in {\alg} directly depends on the prompt. As has been shown in previous work, different prompts have a significant influence on the performance of prompt-based methods~\citep{shin2020autoprompt, gao2021making}. However, in {\alg}, the prompts are fixed and will not be optimized during training.
Therefore, we consider a simple yet effective way to improve the performance through prompt refinement. Specifically, because we automatically generate 100 prompts for each dataset but only use 10 of them, we may select the top 10 prompts following some heuristics to improve the quality of the prompts. Before training, we first evaluate the performance on the validation set by training a weak classifier using the method in Section\,\ref{sec: verbalizer_learning} on the unweighted few-shot training set. Then we construct the prompt pool by selecting the top 10 prompts according to the accuracy of the corresponding weak learner on the validation set. Please note that the few-shot setting makes the refinement process very efficient. Later on, {\alg} is trained using the refined prompts. We mainly evaluate the effectiveness of the prompt refinement on SNLI, MNLI, and QNLI datasets where the gap between {\alg} and standard fine-tuning is relatively large with the increase of training data. Experiment results can be found in Figure\,\ref{fig:datasize}. There are consistent improvements in few-shot performance across three NLI tasks, especially on the QNLI dataset where the performance of {\alg} was far from satisfactory without prompt refinement. Overall, the prompt refinement leads to a trade-off between training cost and model performance.

\noindent \textbf{Effect of the number of prompts} \ 
In our main experiments, we use 10 prompts by default. Intuitively, a large prompt pool increases the diversity of weak classifiers which could improve the performance. However, the training/inference cost will also increase if more prompts are included for model training. We empirically study the relationship between the number of prompts and the model performance in Table\,\ref{tab:template_num} in Appendix\,\ref{sec: addtion_exp}. In general, more prompts benefit the performance in most datasets (except QNLI). We highlight the effectiveness of multiple prompts on AG's News and TREC dataset, on which the performance becomes better and more stable. As we have discussed in the few-shot experiments in Table\,\ref{tab:fewshot_exp}, individual prompt performs very badly on TREC dataset. This is also proved by {\alg}-1 that only achieves 41.3\% accuracy. However, by using our prompt ensemble framework, the performance can be boosted to 84.6\% when 10 prompts are provided. Finally, the performance improvement is relatively small when the number of prompts increases from 10 to 20, implying that 10 prompts should be good enough for {\alg}.

\noindent \textbf{Effect of the prompt generation method} \ 
We also study the performance of {\alg} when combined with other prompt generation methods. Specifically, the following three alternative methods are compared with: (a) {\alg} + LM-BFF with different prompt sets, where we select a different 10-prompt subset from the large number of prompts generated by LM-BFF; (b) {\alg} + prompts from PET~\citep{schick2021exploiting}; and (c) {\alg} + manually written prompts, where we asked several computer science students to write the prompts for each task. All the prompts are listed in the supplemental materials. The results are shown in Table\,\ref{tab:prompt_study}.
\begin{table}[!htp]
    \centering
    \vspace{-2mm}
    \caption{\footnotesize{Performance of {\alg} with different prompt sets. We test the performance with another two different prompt sets generated by LM-BFF~\citep{gao2021making}, one prompt set from PET~\citep{schick2021exploiting}, and one prompt set written by humans (denoted as `LM-BFF set 1', `LM-BFF set 2', `PET' and `Manual' respectively).}}
    \label{tab:prompt_study}
    \vspace{1mm}
    \resizebox{\linewidth}{!}{
    \begin{tabular}{lcc}
    \toprule[1pt]
          & AG's News & RTE \\
    \midrule
          {\alg} (original) & 85.2 (0.9) & 60.0 (5.5) \\
          {\alg} (LM-BFF set 1) & 85.4 (1.6) & 59.1 (5.4) \\
          {\alg} (LM-BFF set 2) & 84.7 (1.3) & 60.3 (6.1) \\
          {\alg} (PET) & 85.0 (1.2) & 58.2 (3.7) \\
          {\alg} (Manual) & 85.2 (0.8) & 60.6 (2.0) \\
    \bottomrule[1pt]
    \end{tabular}
    }
\vspace{-6mm}
\end{table}

{\alg} maintains a consistently competitive performance \emph{regardless of} the prompt generation method used. This further verifies that what truly differentiates our work is the new way to get an efficient and strong black-box classifier which \emph{removes the need to optimize over the prompts}. By shifting the optimization target to the verbalizer compensated by ensembling in an effort, we show that a strong black-box classifier can be obtained without strict requirements on the quality of the prompts.

\noindent \textbf{Ablation studies} and \textbf{Full data training}\ 
We conduct ablation studies on the verbalizer determination method and the prompt ensemble method in Table\,\ref{tab:ablation_ensemble} in Appendix\,\ref{sec: addtion_exp}, showing that both the two modules contribute to the final performance. Also, {\alg} can generalize to full data training instead of just the few-shot setting due to the efficiency. 
We compare {\alg} with fine-tuning on the entire training dataset in Table \ref{tab: full_data} in Appendix\,\ref{sec: addtion_exp}.

\section{Conclusion}
In this paper, we propose {\alg}, an effective black-box model tuning framework. Without access to the parameters and gradients of pre-trained LMs, {\alg} can adapt LMs for various downstream tasks. The efficient weak learner construction method, together with the \textsc{AdaBoost} ensemble algorithm, makes {\alg} achieve state-of-the-art performance in black-box tuning setting with at least 10x run-time efficiency. 

For future directions, we will explore how to generalize {\alg} to more applications, \emph{e.g.}, chain-of-thought prompting~\citep{wei2022chain}. Also, we will study how to combine the prompt ensemble idea in {\alg} with gradient-based optimization and improve the performance of existing prompt-based learning methods.

\section{Acknowledgement}
The work of Bairu Hou and Shiyu Chang was partially supported by National Science Foundation (NSF) Grant IIS-2207052. The computing resources used in this work were partially supported by the MIT-IBM Watson AI Lab.

\bibliography{ref}
\bibliographystyle{icml2023}

\newpage
\appendix
\onecolumn
\section{Implementation Details}
\label{sec: implementation}
\paragraph{Dataset Statistics}
The dataset statistics can be found in Table\,\ref{tab: dataset_stat}. For a fair comparison, the few-shot training/validation/testing split generation is strictly following the implementation of \citet{gao2021making}. 

\begin{wraptable}{r}{.48\textwidth}
    \centering
    \vspace{-8mm}
    \caption{The dataset statistics. $|\mathcal{Y}|$ is the number of classes, Avg.\#W is the average number of words in the input, and \#Train/\#Test refers to the number of examples in the training/testing dataset.}
    \label{tab: dataset_stat}
    \vspace{1mm}
    \resizebox{0.5\textwidth}{!}{
    \begin{tabular}{cccccc}
    \toprule[1pt]
    Category & Dataset & $|\mathcal{Y}|$ & Avg. \#W & \#Train & \#Test \\
    \midrule
    \multirow{4}{*}{single sentence} & SST-2 & 2 & 17 & 6920 & 872 \\
         & SST-5 & 5 & 18 & 8544 & 2210 \\
         & MR & 2 & 20 & 8662 & 2000 \\
         & CR & 2 & 19 & 1775 & 2000 \\
         & AG's News & 4 & 47 & 120000 & 7600 \\
         & TREC & 6 & 10 & 5452 & 500 \\
         & MPQA & 2 & 3 & 8606 & 2000 \\
         & Subj & 2 & 23 & 8000 & 2000 \\
\midrule
    \multirow{4}{*}{sentence pair} &  SNLI & 3 & 22 & 549367 & 9842\\
         & MNLI & 3 & 33 & 392702 & 9815\\
         & QNLI & 3 & 41 & 104743 & 5463\\
         & RTE & 3 & 59 & 2490 & 277\\
         & MRPC & 2 & 43 & 3668 & 408\\
\bottomrule[1pt]
    \end{tabular}
    }
\end{wraptable}
\paragraph{Training of baselines} For standard fine-tuning, we adopt the Huggingface \texttt{transformers} library~\citep{wolf2019huggingface} to load RoBERTa-large backbone model and use its \texttt{Trainer} for fine-tuning. The learning rate is set to 1e-5. We use AdamW optimizer as the optimizer and the learning rate linearly decays to 0. The training batch size is set to 16 and the total training epochs is 100. For the Feature-MLP method, we use a three-layer MLP with a hidden dimension of 100. The learning rate is set to 1e-3 without learning rate decay. We also train the MLP for 100 epochs. For other baselines, we use their official implementation with default hyper-parameters including LM-BFF~\citep{gao2021making}, DART~\citep{zhang2021differentiable}, BBT~\citep{sun2022bbt}, BBTv2~\citep{sun2022bbtv2} and RLPrompt~\citep{deng2022rlprompt}. For RLPrompt, because of its low efficiency, we set its training epochs to 1000 instead of the 12000 used in their paper. This is reasonable since it takes nearly 2 hours for RLPrompt to finish 1000 epochs of optimization.

\section{Additional Experiments}
\label{sec: addtion_exp}

\paragraph{Effect of the number of prompts}
We visualize the relationship between the number of prompts and the model performance in Table\,\ref{tab:template_num}. As we discussed in the main paper, more prompts benefit the performance in most datasets (except QNLI). Also, the performance improvement is relatively small when the number of prompts increases from 10 to 20, implying that 10 prompts should be good enough for {\alg}.

\begin{table*}[h]
    \centering
    \vspace{-3mm}
    \caption{\footnotesize{Performance of {\alg} with different numbers of prompts in few-shot setting ($k = 16$). {\alg}-$d$ means top-$d$ prompts (sorted according to the beam search score) are used for model training. Mean accuracy (and standard deviation) is reported over 5 different splits.}}
    \label{tab:template_num}
    \vspace{1mm}
    \resizebox{0.95\textwidth}{!}{
    \begin{tabular}{lccccccccc}
    \toprule[1pt]
          & SST-2 & MR & AG's News & TREC & SNLI & MNLI & QNLI & RTE & Avg.  \\
    \midrule
         {\alg}-1 & 86.1 (1.0) & 85.1 (5.0) & 73.3 (3.7) & 41.3 (4.3) & 53.4 (4.0) & 49.5 (3.5) & 58.0 (2.4)& 56.5 (5.7) &  62.9\\
         {\alg}-5 & 88.8 (1.9) & 87.9 (1.6) & 83.5 (4.2) & 78.0 (2.5)& 59.1 (3.5) & 50.9 (4.8) & 56.5 (2.1) & 57.0 (4.5) & 70.2\\
         {\alg}-10& 87.6 (3.0) & 84.6 (2.5) & 85.2 (0.9) & 81.6 (4.0) & 61.3 (3.5)& 52.5 (1.5) & 58.0 (3.3)& 60.0 (5.5) &  71.4 \\
         {\alg}-20 & 88.1 (2.6)& 84.0 (2.3) & 86.4 (1.3)& 81.9 (2.7) & 60.8 (3.9) & 55.2 (1.2) & 57.0 (4.4) & 57.1 (3.2)&  71.3\\
    \bottomrule[1pt]
    \end{tabular}
    }
\end{table*}

\paragraph{Effect of the verbalizer construction and the prompt ensemble method}
We conduct an ablation study to demonstrate the effectiveness of the proposed verbalizer construction method. Furthermore, since {\alg} use \textsc{AdaBoost} algorithm to ensemble the weak learners, we also study the performance of our method when using other prompt ensemble methods. Specifically, we design the following baselines. (a) \textbf{Ensemble (rand)}. We pair each prompt with a randomly generated verbalizer (randomly select a token from the vocabulary for each class). Then the 10 weak learners are ensembled using majority vote (b) \textbf{Ensemble (manual)}. Instead of using random verbalizers, we take the manually designed verbalizers from LM-BFF~\citep{gao2021making}. Then the 10 prompts will be paired with the manual verbalizer to form 10 weak learners that are ensembled using majority vote (c) \textbf{\textsc{PromptVoting}}. We use our verbalizer construction method to find the verbalizer for each prompt (on the unweighted 16-shot training dataset). Then we ensemble the 10 weak classifiers using the majority vote. The experiment results are shown in Table\,\ref{tab:ablation_ensemble}.
\begin{table*}[h]
    \centering
    \vspace{-3mm}
    \caption{\footnotesize{Ablation study of the verbalizer construction and the prompt ensemble method in {\alg} in a few-shot setting ($k=16$) measured by classification accuracy (\%). All methods use RoBERTa-large~\citep{liu2019roberta} as the backbone LM for a fair comparison. The best results from black-box methods are highlighted in \textbf{bold}.
    }}
    \label{tab:ablation_ensemble}
    \vspace{1mm}
    \resizebox{0.9\textwidth}{!}{
    \begin{tabular}{l|cccccccc}
    \toprule[1pt]
         Method & SST-2 & MR & AG's News & TREC & SNLI & MNLI & QNLI & RTE  \\
    \midrule
         Fine-tuning & 81.4 (3.8)& 82.7 (3.6)& 86.2 (1.4) & 88.8 (2.1) & 48.4 (4.8) & 45.8 (6.4) & 56.3 (1.5) & 54.4 (3.9)\\
         Best baseline & 90.5 (1.5) & 86.2 (2.5) & 83.6 (2.0) & 63.8 (9.9)& 57.4 (2.7) & 51.4 (3.3) & \textbf{58.1 (2.5)} & 53.2 (7.0)\\
         Ensemble (rand) & 50.9 & 50.0 & 25.2 & 17.4 & 36.5 & 36.0 & 49.7 & 47.3 \\
         Ensemble (manual) & 85.8 & 83.9 & 67.1 & 31.6 & 41.3 & 47.4 & 55.1 & 50.2 \\
\rowcolor{Gray}
         \textsc{PromptVoting} & \textbf{91.2 (1.2)} & \textbf{87.2 (0.9)} & 83.5 (1.3) & 67.2 (6.7) & 55.6 (3.0)& 48.9 (2.9) & 56.2 (3.0) & 54.6 (1.4) \\
\rowcolor{Gray}
         {\alg} & 87.6 (3.0) & 84.6 (2.5) & \textbf{85.2 (0.9)} & \textbf{81.6 (4.0)} & \textbf{61.3 (3.5)}& \textbf{52.5 (1.5)} & 58.0 (3.3) & \textbf{60.0 (5.5)} \\
    \bottomrule[1pt]
    \end{tabular}
    }
\end{table*}

From the experiment results above, we highlight the following conclusions. (a) \textbf{The effectiveness of our verbalizer construction method.} Given the same ensemble scheme (majority voting with 10 prompts), \textsc{PromptVoting} consistently outperforms Ensemble (rand) and Ensemble (manual), indicating the effectiveness of our verbalizer construction method. It is worth noting that even on sentiment classification datasets (SST-2 and MR) which are very intuitive to construct verbalizers, \textsc{PromptVoting} is still largely better than manually defined verbalizers. \textsc{PromptVoting} even outperforms state-of-the-art methods on some datasets. (b) \textbf{The More advanced ensemble improves the performance.} One can clearly observe improvement when we change the majority vote to the Adaboost ensemble. It is also worth mentioning that PromptVoting offers an alternative solution to ensemble weak learners on SST-2 and MR datasets where \textsc{Adaboost} cannot be used\footnote{We discussed why \textsc{Adaboost} cannot be used on SST-2 and MR datasets in Section\,\ref{subsec: eval_result}. Individual weak learners can achieve 100\% accuracy on the training dataset and \textsc{Adaboost} cannot be used to ensemble models that achieve 100\% accuracy.}. 
In a summary, each module in our method contributes to the final performance.

\paragraph{Improve the efficiency of {\alg}} 
In Table\,\ref{tab:efficiency} in the main paper, we visualize the time cost of different algorithms and demonstrate that {\alg} improves the efficiency over existing black-box baselines by more than 10 times. In fact, the training efficiency of {\alg} can be further improved by adjusting the hyper-parameters.

Specifically,  recall that when we screen the verbalizer, we will take the top-$m$ tokens for each class and use brute force to find the best one. For AG's News, $m$ is set to 10 and for RTE,$m$ is set to 50. That is, for each weak learner, there are $10^4$ verbalizers and $50^2$ for AG's News and RTE respectively. The main reasons that we use such a large $m$ are two-fold. Firstly, we hope we can trade off more time for better weak classifier performance. Secondly, the efficiency of our baselines is very low. Even though we use a large $k$ and spend much time on constructing individual classifiers, our method is still 10x faster than existing black-box baselines. Therefore, we did not use a smaller $k$ for better efficiency in the main experiments. 

Ideally, we can use a smaller k (i.e., $m$ = 5 for AG's News and $m$ = 10 for RTE) which can largely improve the efficiency without hurting the performance. The comparison between our original settings and the new settings with a smaller $m$ is shown in Table\,\ref{tab:improve_efficiency}. One can clearly observe that the {\alg} can achieve the best efficiency (2 minutes and 0.7 minutes for training on AG's News and RTE datasets respectively).

\begin{table*}[h]
    \centering
    \caption{\footnotesize{Deployment efficiency of proposed {\alg} and baseline methods in few-shot setting ($k=16$). Wall time is reported to measure the training time efficiency. Query efficiency is evaluated by \#Forward and \#Backward, which refer to the number of forward/backward passes per batch during training respectively. We include another variant of {\alg} with a smaller $m$ ($m$ = 5 for AG's News and $m$ = 10 for RTE) when screening the verbalizer.}}
    \vspace{1mm}
    \label{tab:improve_efficiency}
    \resizebox{0.95\textwidth}{!}{
    \begin{tabular}{l|cc|cccc|cccc}
    \toprule[1pt]
         \multirow{2}{*}{Method} & \multirow{2}{*}{\makecell[c]{Trainable\\param}} & \multirow{2}{*}{\makecell[c]{Total\\param}} & \multicolumn{4}{c|}{AG's News} & \multicolumn{4}{c}{RTE}  \\
         & & & Acc & Wall Time & \#Forward &\#Backward & Acc & Wall Time & \#Forward &\#Backward \\
    \midrule
         Fine-tuning & 335M & 335M & 86.2 & 13 min & 100 & 100 & 54.4 & 19 min & 100 & 100 \\
         LM-BFF~\citep{gao2021making} & 335M & 335M & 87.1 & 5min & 32 & 32 & 66.6 & 9 min & 60 & 60\\
         DART~\citep{zhang2021differentiable} & 335M & 335M & 86.8 & 15 min & 30 & 30 & 59.0 & 5 min & 120 & 120 \\
    \midrule
         BBT~\citep{sun2022bbt} & 25k & 335M & 81.2 & 88 min& 8K & 0 & 49.1 & 52 min & 8K & 0 \\
         BBTv2~\citep{sun2022bbtv2} & 25k & 335M & 83.6 & 90 min & 8K & 0 & 53.2 & 70 min & 8K & 0 \\
         RLPrompt~\citep{deng2022rlprompt} & 3M & 420M & 77.2 & 117 min & 1K & 0 & 52.2 & 90 min & 1K & 0 \\
\rowcolor{Gray}
         {\alg} & $<$1k & 335M & 85.2 & 8 min & 10 & 0 & 60.0 & 4 min & 10 & 0\\
\rowcolor{Gray}
         {\alg} (small $k$) & $<$1k & 335M & 84.4 & \textbf{2 min} & 10 & 0 & 59.1 & \textbf{0.7 min} & 10 & 0\\

    \bottomrule[1pt]
    \end{tabular}
    }
\end{table*}

\paragraph{Performance on full dataset} The high efficiency of {\alg} makes it possible to generalize to medium-sized datasets. 
We evaluate the performance of {\alg} on SST-2, MR, TREC, and RTE datasets. We sample 10\% of the original training set to construct the validation set and use the original validation set for testing if the labeled test set is unavailable. 
The experiment results can be found in Table \,\ref{tab: full_data}. 
\begin{wraptable}{r}{.48\textwidth}
    \centering
    \vspace{-3mm}
    \caption{Performance of full data training}
    \vspace{1mm}
    \label{tab: full_data}
    \resizebox{0.5\textwidth}{!}{
    \begin{tabular}{ccccc}
    \toprule[1pt]
     & SST-2 & MR & TREC & RTE \\
    \midrule
    Fine-tuning & 95.5 (0.4) & 91.5 (0.6) & 97.2 (0.2) & 81.9 (1.1)\\
    {\alg} & 94.1 (0.3) & 89.7 (0.4) & 90.5 (1.2) & 71.7 (2.0)\\
    \bottomrule[1pt]
    \end{tabular}
    }
\end{wraptable}
{\alg} achieves comparable performance with standard fine-tuning on SST-2 and MR datasets, which is impressive given the fact that {\alg} has no access to the parameters and gradients of the LM. For the TREC dataset, standard fine-tuning outperforms {\alg}, but we still remark that the performance is still highly usable in the black-box setting. Finally, the gap between {\alg} and fine-tuning is relatively large on the RTE dataset, which is consistent with our previous discovery that it seems pre-trained LMs are not good at sentence pair classification tasks before fine-tuning.

\paragraph{Experiments on more datasets} 
In Table\,\ref{tab:all_dataset} we display the experiments on all datasets. Similar to SST-2 and MR, we do not ensemble weak learners on the CR dataset since even individual weak learners in {\alg} can achieve 100\% accuracy on the training set. Instead, we directly report the performance of the individual weak learner that performs best on the validation set. According to the experiments, our conclusion still holds that {\alg} achieves state-of-the-art performance on a wide range of datasets. 
\begin{table}[h]
    \centering
    \caption{\footnotesize{Performance of proposed {\alg} and baseline methods in few-shot setting ($k=16$) measured by classification accuracy (\%) and F1 score (for the MRPC dataset only). All methods use RoBERTa-large~\citep{liu2019roberta} as the backbone LM for a fair comparison. 
    Two white-box methods are included for reference including LM-BFF~\citep{gao2021making} and DART~\citep{zhang2021differentiable}. Feature-MLP, BBT~\citep{sun2022bbt}, BBTv2~\citep{sun2022bbtv2} and RLPrompt~\citep{deng2022rlprompt} are the main black-box baselines. {\alg}-32 combines both training and validation sets for training.
    Mean accuracy (and standard deviation) is reported over 5 different splits. The best results are highlighted in \textbf{bold} and the second best are \underline{underlined}.}}
    \label{tab:all_dataset}
    \vspace{1mm}
    \resizebox{0.8\textwidth}{!}{
    \begin{tabular}{l|ccccccc}
    \toprule[1pt]
         Method & SST-2 & MR & AG's News & TREC & SNLI & MNLI & QNLI  \\
    \midrule
         Fine-tuning & 81.4 (3.8)& 82.7 (3.6)& 86.2 (1.4) & 88.8 (2.1) & 48.4 (4.8) & 45.8 (6.4) & 56.3 (1.5) \\
         LM-BFF & 92.3 (1.5)& 87.4 (0.6) & 87.1 (1.2) & 83.4 (2.7) & 76.5 (2.6)& 68.7 (2.0) & 64.4 (4.6) \\
         DART & 93.5 (0.5) & 88.2 (1.0) & 86.8 (0.5) & 87.1 (3.8)& 75.8 (1.6) & 67.5 (2.6) & 66.7 (3.7) \\
    \midrule
         BBT & \underline{88.2 (1.7)} & 82.8 (2.6) & 81.2 (2.7) & 39.3 (5.2)& 44.7 (4.0)& 42.3 (2.8) & 56.8 (2.0) \\
         BBTv2& \underline{88.5 (2.1)} & 83.7 (1.8) & 83.6 (2.0) & 63.8 (9.9) & 57.4 (2.7) & 51.4 (3.3) & \underline{58.1 (2.5)}  \\
         RLPrompt & \textbf{90.5 (1.5)} & \textbf{86.2 (2.5)} & 76.2 (2.7)& 37.3 (3.5)& 42.9 (1.8) & 40.7 (4.7)& 52.1 (2.9) \\
\rowcolor{Gray}
         {\alg} & 87.6 (3.0) & \underline{84.6 (2.5)} & \textbf{85.2 (0.9)} & \underline{81.6 (4.0)} & \underline{61.3 (3.5)}& \underline{52.5 (1.5)} & \underline{58.0 (3.3)} \\
\rowcolor{Gray}
         {\alg}-32 & 87.6 (3.3) & 84.7 (2.1) & \underline{84.2 (1.1)} & \textbf{84.5 (1.4)} &  \textbf{62.0 (2.7)} & \textbf{53.8 (1.2)} & \textbf{58.3 (2.8)} \\
    \midrule
    & & &  & & & &  \\
    \midrule
         Method & RTE & SST-5 & CR & MPQA & Subj  & MRPC &  Avg.  \\
    \midrule
         Fine-tuning& 54.4 (3.9) & 43.9 (2.0) & 75.8 (3.2) & 72.0 (3.8) & 90.8 (1.8) & 76.6 (2.5) & 69.5\\
         LM-BFF& 66.6 (6.4) & 48.5 (1.5) & 89.2 (3.8) & 83.7 (2.4) & 90.7 (1.9) & 76.0 (3.4) & 78.0 \\
         DART& 59.0 (2.5) & 48.6 (1.5) &  91.8 (0.5) & 68.1 (8.9) & 90.7 (1.4) & 78.3 (4.5) & 77.1\\
         \midrule
         BBT & 49.1 (3.3) & 36.3 (3.6) & 86.2 (1.3) & \underline{78.4 (2.2)} & 75.6 (3.2) & \textbf{73.7 (6.0)} & 64.2 \\
         BBTv2 & 53.2 (7.0) & 38.7 (2.2) & \textbf{88.5 (1.0)} & \textbf{80.6 (2.7)} & 78.2 (2.9) & \underline{73.6 (7.6)} & 69.2  \\
         RLPrompt & 52.2 (2.2) & 40.1 (1.9) & 87.4 (1.7) & 69.4 (3.7) & 81.9 (1.2) & 61.9 (5.1) & 63.0 \\
\rowcolor{Gray}
         {\alg}& \underline{60.0 (5.5)} & \underline{42.3 (1.8)} & 86.8 (0.8) & 72.7 (3.4) & \underline{86.1 (4.8)} & 70.5 (2.9) & \underline{71.5} \\
\rowcolor{Gray}
         {\alg}-32 & \textbf{60.3 (2.4)} & \textbf{44.0 (1.5)} & \underline{88.1 (1.1)} & 75.4 (2.3) & \textbf{90.4 (1.1)} & 69.2 (5.8) & \textbf{72.5}\\
    \bottomrule[1pt]
    \end{tabular}
    }
\end{table}

\begin{figure}[h]
    \centering
    \subfigure[Performance on AG]
    {          
        \includegraphics[width=.23\textwidth, height=!]{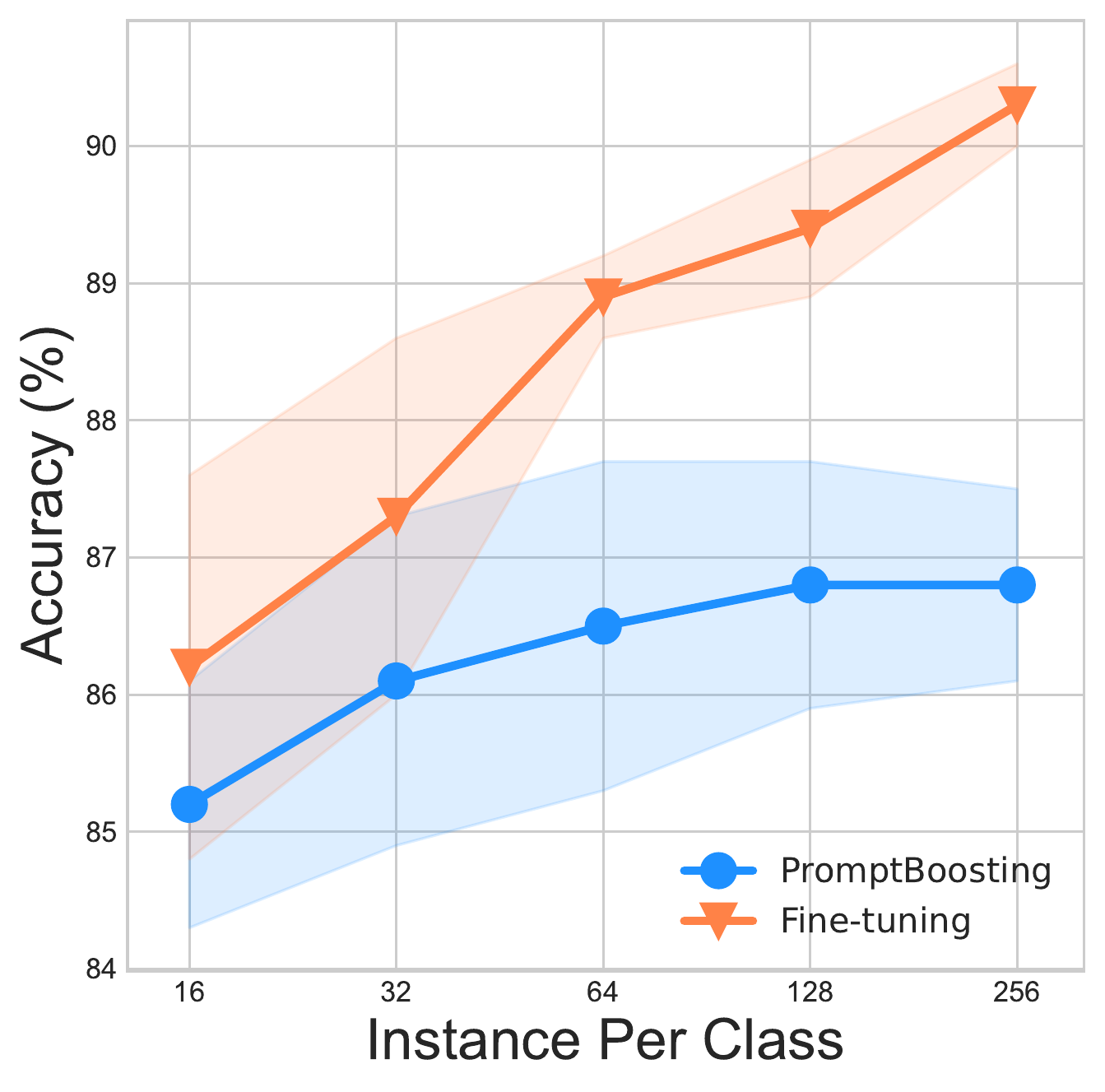}
    }
    \hspace{-1mm}
    \subfigure[Performance on TREC]
    {          
        \includegraphics[width=.23\textwidth, height=!]{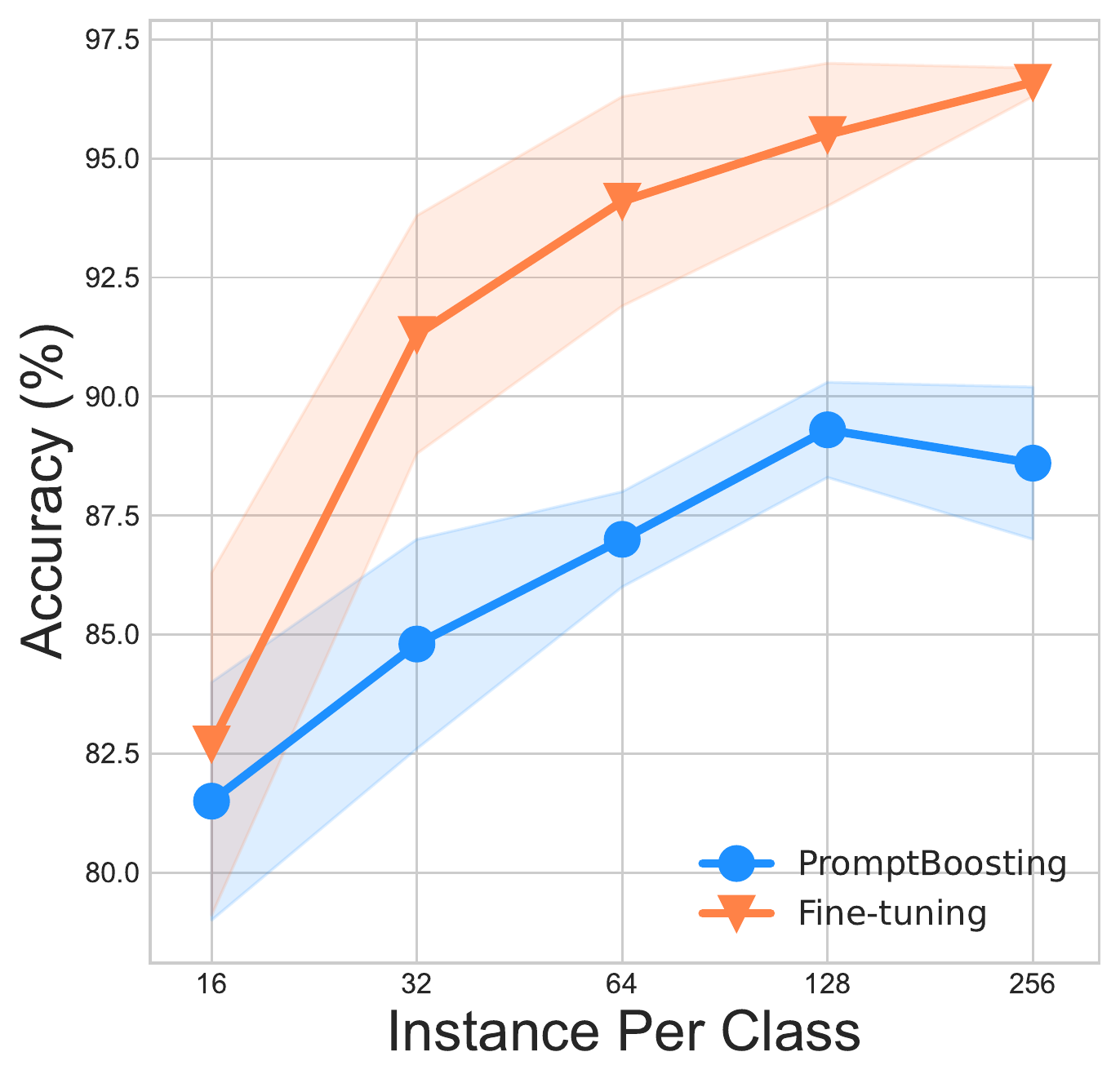}
    }
    \hspace{-1mm}
    \subfigure[Performance on QNLI]
    {          
        \includegraphics[width=.23\textwidth, height=!]{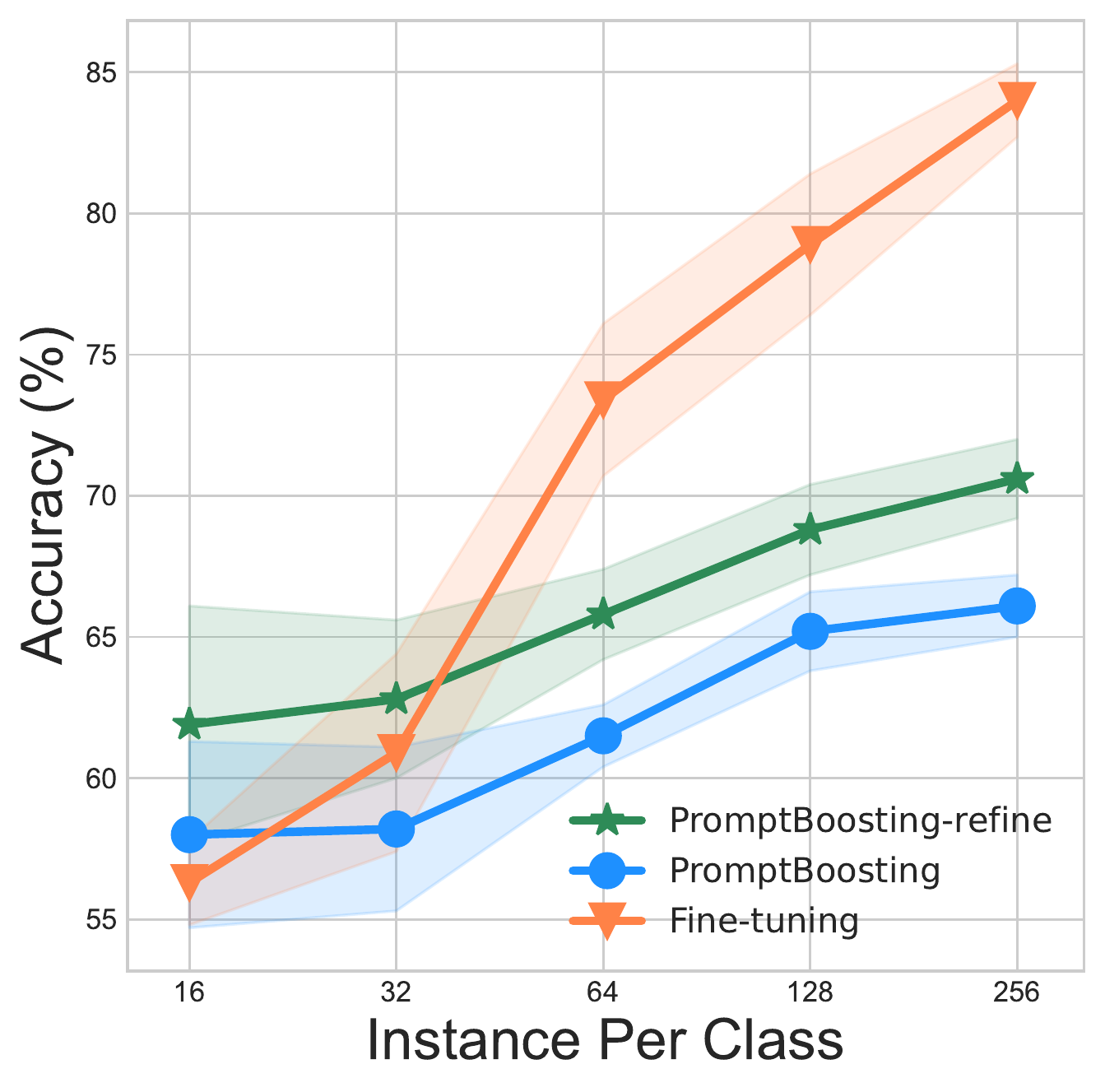}
    }
    \hspace{-1mm}
    \subfigure[Performance on RTE]
    {          
        \includegraphics[width=.23\textwidth, height=!]{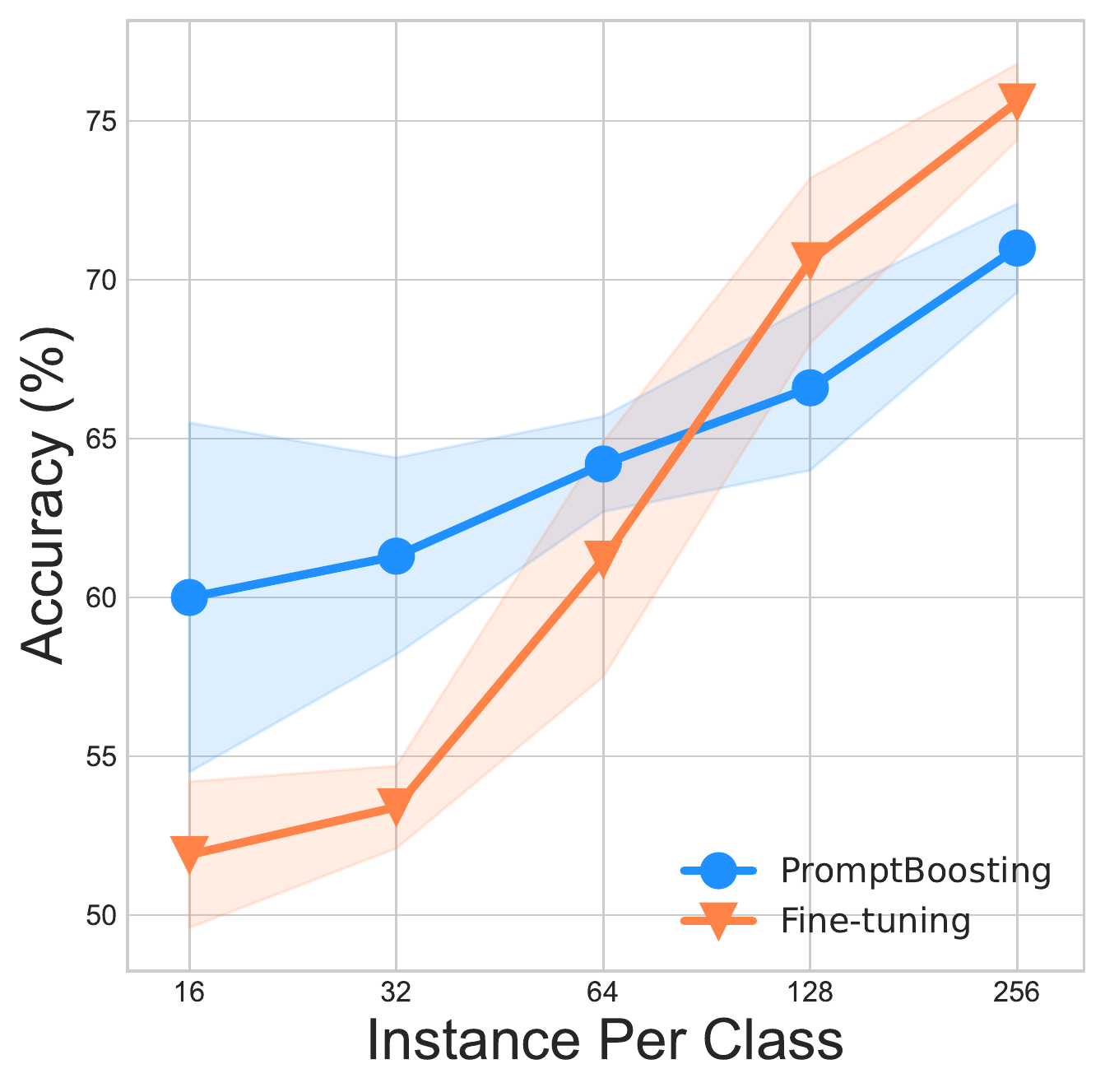}
    }

    \caption{Model performance as a function of training set size on different datasets. For the QNLI dataset, we also include prompt refinement for better performance.}
\label{fig: appendix_training_size}
\vspace{-0.5em}
\end{figure}

\paragraph{Effect of training data size} For AGNews, TREC, QNLI, and RTE datasets, we show the performance of {\alg} as the size of the training set increases in Figure\,\ref{fig: appendix_training_size}.

\section{Generated Prompts}
\label{sec: templates_visualize}
In this section, we visualize the prompts we used in our experiments for each dataset in Table \,\ref{tab: template_visualize}. Regardless of different few-shot training/validation splits, we use the same 10 prompts for model training.

\begin{table}[t]
    \centering
    \caption{Prompts used by {\alg} on different datasets.}
    \label{tab: template_visualize}
    \resizebox{0.95\textwidth}{!}{
    \begin{tabular}{c|l|l}
    \toprule[1pt]
        & \quad \quad \quad \quad \quad \quad \quad \quad SST-2 & \quad \quad \quad \quad \quad \quad \quad \quad MR \\
    \midrule
        1 &\texttt{[Input] It's [MASK].} &  \texttt{[Input] It's [MASK].}\\
        2 &\texttt{[Input] A [MASK] movie.} & \texttt{[Input] It's [MASK]!} \\
        3 &\texttt{[Input] A [MASK] film.} & \texttt{[Input] A [MASK] piece of work.}\\
        4 &\texttt{[Input] A [MASK] piece of work.} & \texttt{[Input] It’s [MASK].} \\
        5 &\texttt{[Input] A truly [MASK] film.} & \texttt{[Input] A [MASK] waste of time.} \\
        6 &\texttt{[Input] This is [MASK].} & \texttt{[Input] A truly [MASK] film.} \\
        7 &\texttt{[Input] It was [MASK].} & \texttt{[Input] I thought it was [MASK].} \\
        8 &\texttt{[Input] A [MASK] waste of time.} & \texttt{[Input] It's just [MASK].} \\
        9 &\texttt{[Input] It's [MASK]!} & \texttt{[Input] A truly [MASK] movie.}  \\
        10 &\texttt{[Input] A truly [MASK] movie.} & \texttt{[Input] The film is [MASK].}  \\
    \midrule
        &  \quad \quad \quad \quad \quad \quad \quad \quad \quad \quad AG's News & \quad \quad \quad \quad \quad \quad \quad \quad TREC \\
    \midrule
        1 &\texttt{[Input] This entry was posted in [MASK].} &  \texttt{[Input] What is [MASK]?}\\
        2 &\texttt{[Input] U.S. [MASK] News.} & \texttt{[Input] What is the [MASK]?} \\
        3 &\texttt{[Input] U.S. [MASK].} & \texttt{[Input] What [MASK]?}\\
        4 &\texttt{[Input]  This entry was posted in [MASK] News.} & \texttt{[Input] The [MASK].} \\
        5 &\texttt{[Input] The [MASK] Journal reports.} & \texttt{[Input] See [MASK].} \\
        6 &\texttt{[Input] The [MASK] Journal has more.} & \texttt{[Input] Which [MASK]?} \\
        7 &\texttt{[Input] Read more at [MASK] News Now.} & \texttt{[Input] The [MASK]?} \\
        8 &\texttt{[Input] The New York Times [MASK].} & \texttt{[Input] Full [MASK].} \\
        9 &\texttt{[Input] The New York Times [MASK] Report.} & \texttt{[Input] How many [MASK]?}  \\
        10 &\texttt{[Input] Read more at[MASK] Insider.} & \texttt{[Input] 1.[MASK].}  \\
    \midrule
        & \quad \quad \quad \quad \quad \quad \quad \quad SNLI & \quad \quad \quad \quad \quad \quad \quad \quad MNLI \\
    \midrule
        1 &\texttt{[Input1]. [MASK], [Input2]} &  \texttt{[Input1]. [MASK], [Input2]}\\
        2 &\texttt{[Input1]. [MASK]. [Input2]}& \texttt{[Input1]. [MASK], but [Input2]}\\
        3 &\texttt{[Input1]. [MASK] and [Input2]} & \texttt{[Input1]. [MASK]. [Input2]}\\
        4 &\texttt{[Input1]. [MASK], but [Input2]} & \texttt{[Input1]! [MASK], [Input2]} \\
        5 &\texttt{[Input1]. [MASK]: [Input2]} & \texttt{[Input1]. [MASK]. But [Input2]} \\
        6 &\texttt{[Input1]. [MASK] one of [Input2]}& \texttt{[Input1]? [MASK], [Input2]} \\
        7 &\texttt{[Input1]. [MASK]... [Input2]}& \texttt{[Input1]. [MASK] and [Input2]}\\
        8 &\texttt{[Input1]. [MASK], just [Input2]} & \texttt{[Input1]. [MASK], and [Input2]} \\
        9 &\texttt{[Input1]. [MASK] it is [Input2]} &\texttt{[Input1]. [MASK] but [Input2]} \\
        10 &\texttt{[Input1]. [MASK]; [Input2]} &\texttt{[Input1]. [MASK]... [Input2]} \\
    \midrule
        & \quad \quad \quad \quad \quad \quad \quad \quad QNLI & \quad \quad \quad \quad \quad \quad \quad \quad RTE \\
    \midrule
        1 & \texttt{[Input1]? [MASK], [Input2]} &  \texttt{[Input1]. [MASK], [Input2]} \\
        2 & \texttt{[Input1]? [MASK], but [Input2]} & \texttt{[Input1]. [MASK]. [Input2]}\\
        3 & \texttt{[Input1]? [MASK]. [Input2]} & \texttt{[Input1]. [MASK], but [Input2]}\\
        4 & \texttt{[Input1]? [MASK]. But [Input2]}  & \texttt{[Input1]. [MASK] and [Input2]} \\
        5 & \texttt{[Input1]? [MASK]. In fact, [Input2]} & \texttt{[Input1]. [MASK]: [Input2]} \\
        6 & \texttt{[Input1]? [MASK]; [Input2]} & \texttt{[Input1]. [MASK], the [Input2]}\\
        7 & \texttt{[Input1]? [MASK]. However, [Input2]} & \texttt{[Input1]. [MASK]; [Input2]}\\
        8 & \texttt{[Input1]? [MASK], and [Input2]} & \texttt{[Input1]. [MASK]-[Input2]} \\
        9 & \texttt{[Input1]? [MASK]: [Input2]}  &\texttt{[Input1]. [MASK], and [Input2]} \\
        10 & \texttt{[Input1]. [MASK], [Input2]} &\texttt{[Input1]. [MASK] but [Input2]} \\
    \bottomrule[1pt]
    \end{tabular}
    }
\end{table}

\end{document}